# A Novel Hybrid Sampling Framework for Imbalanced Learning

Asif Newaz [1]*, Farhan Shahriyar Haq [2]

[1] Department of Electrical and Electronic Engineering,

Islamic University of Technology, Gazipur, Bangladesh;

[2] Robi Axiata Limited, Dhaka, Bangladesh;

**Email:** eee.asifnewaz@iut-dhaka.edu [1], farhanshahriyar@iut-dhaka.edu [2]

*** Corresponding Author:**

**Address:** Department of Electrical and Electronic Engineering, Islamic University of Technology, Gazipur, Bangladesh - 1704.

**Email:** eee.asifnewaz@iut-dhaka.edu

**Contact:** +8801880841119

**Abstract**

Class imbalance is a frequently occurring scenario in classification tasks. Learning from imbalanced data poses a major challenge, which has instigated a lot of research in this area. Data preprocessing using sampling techniques is a standard approach to deal with the imbalance present in the data. Since standard classification algorithms do not perform well on imbalanced data, the dataset needs to be adequately balanced before training. This can be accomplished by oversampling the minority class or undersampling the majority class. In this study, a novel hybrid sampling algorithm has been proposed. To overcome the limitations of the sampling techniques while ensuring the quality of the retained sampled dataset, a sophisticated framework has been developed to properly combine three different sampling techniques. Neighborhood Cleaning rule is first applied to reduce the imbalance. Random undersampling is then strategically coupled with the SMOTE algorithm to obtain an optimal balance in the dataset. This proposed hybrid methodology, termed "SMOTE-RUS-NC", has been compared with other state-of-the-art sampling techniques. The strategy is further incorporated into the ensemble learning framework to obtain a more robust classification algorithm, termed "SRN-BRF". Rigorous experimentation has been conducted on 26 imbalanced datasets with varying degrees of imbalance. In virtually all datasets, the proposed two algorithms outperformed existing sampling strategies, in many cases by a substantial margin. Especially in highly imbalanced datasets where popular sampling techniques failed utterly, they achieved unparalleled performance. The superior results obtained demonstrate the efficacy of the proposed models and their potential to be powerful sampling algorithms in imbalanced domain.

# 1. Introduction

Learning from imbalanced data is a major challenge in classification tasks. Real-world datasets often come with different degrees of imbalance. Standard classification algorithms are designed in such a way that they are trained to minimize the number of misclassifications, irrespective of the class. Therefore, if one class is underrepresented in the data, the performance gets biased towards the majority class. The problem intensifies if the disparity between the classes is larger. The classifier might completely overlook the minority class samples and classify all test samples as the majority class. However, identifying the minority class samples correctly is often imperative. Therefore, necessary steps need to be adopted to address the class imbalance to obtain satisfactory performance. This has attracted a lot of attention from researchers over the years, and several methods have been proposed to tackle the problem.

The standard approach to deal with such an imbalance is to make use of sampling techniques to balance the dataset. Preprocessing the dataset this way before feeding it to the classifiers improves the overall performance. However, the effectiveness of these techniques depends significantly on how the data is resampled. Many different sampling techniques have been proposed by researchers, which can be broadly classified into two groups: undersampling and oversampling. Undersampling refers to eliminating samples from the majority class, while oversampling refers to generating new minority class samples using the existing ones. These can be achieved heuristically or non-heuristically. While non-heuristic approaches are simple and fast, they can cause overfitting or loss of information. Heuristic approaches, on the other hand, are aimed at ensuring the quality of the resampled dataset by strategically generating new minority class samples or carefully removing majority class samples from the original data.

Random oversampling (ROS) and random undersampling (RUS) are two non-heuristic approaches. ROS just creates duplicates of the already existing samples to balance the dataset. This often causes overfitting, leading to an inferior performance on the test data, hence usually less used. RUS algorithm, on the other hand, removes samples from the majority class to attain balance. This is done randomly, so there is a chance of removing potentially important information from the data. To avoid such problems, heuristic methods are employed. SMOTE (Synthetic Minority Oversampling Technique) is one of the most renowned approaches for oversampling [1]. It generates synthetic samples using interpolation. Due to its simplicity and robustness, it has become a standard benchmark for learning from imbalanced data [2]. However, loss of generalization might occur when too many samples are synthesized using SMOTE to balance the data. Different extensions of the SMOTE have also been proposed to overcome its limitations; these include ADASYN (Adaptive Synthetic Oversampling) [3], Borderline-SMOTE [4], DBSMOTE (Density-Based SMOTE) [5], Safe-level SMOTE [6], etc. Undersampling techniques include Tomek Links (TL) [7],

Condensed Nearest Neighbor Rule (CNN) [8], One-Sided Selection (OSS) [9], Near Miss [10] approaches, etc. Each of these methods has its own way of identifying noisy, borderline, or redundant majority class samples for removal. For instance, in TL-based undersampling, when a tomek-link is found, the sample belonging to the majority class is removed. CNN eliminates the majority class samples that are far away from the decision border, considering them irrelevant for learning. OSS is a combination of TL and CNN-based undersampling approaches. It first removes the borderline majority class samples based on Tomek-links, then removes the distant samples using CNN. One prominent issue here is that these heuristic undersampling techniques do not balance the data. The number of majority class samples that are removed depends on the distribution of the data. Therefore, even after performing these undersampling techniques, the resampled data can still remain quite imbalanced. To obtain optimal balance, hybridization of undersampling and oversampling has also been proposed by some researchers [11]. For example, SMOTE-Tomek [12] and SMOTE-ENN [13]. SMOTE-Tomek is just a mere combination of TL with SMOTE, while SMOTE-ENN is a combination of CNN with SMOTE. Since TL or ENN cannot balance the dataset by themselves, first undersampling the data to reduce the imbalance ratio, then performing SMOTE to balance the dataset works comparatively better than using them unilaterally. Several other sampling techniques have been proposed by researchers over the years. Table 1 provides an overview of some of those techniques.

Learning from imbalanced data is often complicated, and the sampling techniques discussed above do not always perform well. It depends on many factors, like the imbalance ratio, class overlapping, or distribution of the data samples. SMOTE works well for low-dimensional data but is less effective when the data is high-dimensional [14]. RUS is found to perform better in high-dimensional data [14]. However, when there is a large imbalance present between the classes, a significant portion of the majority class samples needs to be eliminated to balance the dataset using RUS. This greatly affects the performance of the classifiers. Similarly, generating a large number of synthetic samples using SMOTE causes the model to overfit, resulting in poor generalization. Consequently, SMOTE generally provides comparable performance when the imbalance ratio is small but fails to perform well when a large disparity is present between the classes. A proper hybridization of the sampling techniques can substantially mitigate these problems.

In this study, we propose a hybrid sampling framework to obtain an optimal balance between the classes and to overcome the limitations of the above-mentioned methods. The goal is to combine the sampling techniques in such a way that loss of information can be avoided while achieving well-generalized performance on a wide range of imbalance scenarios. In order to accomplish that, we start by utilizing the Neighborhood Cleaning Rule (NC) algorithm to identify the noisy and ambiguous majority class samples in the data and remove them. NCL is a heuristic undersampling strategy that modifies the ENN algorithm for enhanced data cleaning [15]. Using this approach first reduces the imbalance ratio. Then RUS algorithm is applied to randomly remove some more samples from the majority class. This further lowers the imbalance ratio. Unlike the standard RUS algorithms, only a small portion of the majority class samples are removed in this step. Finally, SMOTE algorithm is utilized to generate some synthetic samples to obtain an optimal balance between the two classes in the data. Since the imbalance ratio has already been truncated due to the combined NCL and RUS approaches, the number of minority class samples that are created using the SMOTE algorithm is limited. Using these approaches unilaterally fails to provide desirable performance as RUS removes a significant number of samples from the data to obtain balance, while SMOTE generates that many samples to bring balance, leading to overfitting. First reducing the disproportion using a heuristic undersampling method like NCL, then merging RUS with SMOTE can provide an optimal balance. This alleviates the problem of information loss that occurs in RUS while also mitigating the problem of loss of generalization caused by SMOTE. This way, the proposed hybrid sampling technique SMOTE-RUS-NC ensures the quality of the resampled data without getting biased towards any particular class and is capable of providing well-generalized performance over a wide range of imbalanced datasets.

Ensemble algorithms like bagging or boosting are quite popular as they tend to outperform traditional machine learning classification algorithms like K-Nearest Neighbors (KNN), Support Vector Machine (SVM), or Decision Trees (DT). However, they still remain susceptible to skewed class distribution. The sampling techniques can be incorporated into the ensemble learning framework for better generalization and performance. Balanced Random Forest (BRF), RUSBoost, SMOTE-Boost, Balanced Bagging, and Easy Ensemble are some of the examples of ensemble methods for imbalanced learning. They incorporate sampling techniques like RUS, ROS, or SMOTE to balance the datasets before training the classifiers. Sampling techniques like RUS is a high-variant strategy. Since samples are eliminated from the data randomly, classifiers trained on such data which are balanced this way tend to provide inconsistent results. The performance may vary significantly over several iterations. This is quite undesirable for real-world applications, making the RUS approach impractical despite its better performance. To alleviate the problem, the RUS algorithm is used in conjunction with the ensemble framework, where a large number of base learners are used. Each base learner is trained on a different subset of the data balanced using the RUS algorithm. The predictions of different base learners are then aggregated to obtain the result. This lowers the variation and mitigates the information loss caused by the RUS methodology. In this study, we integrate our proposed hybrid sampling method into the bagging ensemble framework. Each bootstrap sample is balanced using the proposed technique. The DTs are then trained on different balanced bootstrap samples. The Random Forest (RF) architecture is used to form the ensemble and aggregate the results. We termed this ensemble algorithm as SMOTE-RUS-NC-BRF, or SRN-BRF in short. Combining the hybrid sampling technique with the ensemble framework provides better generalization with improved performance.

Rigorous experiments have been carried out to assess the performance of the proposed approaches. 26 publicly available datasets with different degrees of imbalance (1.8 - 130) are utilized for validation. The results from the data preprocessing approach SMOTE-RUS-NC are compared with other popular sampling techniques like SMOTE, ADASYN, RUS, CNN, NC, SMOTE-ENN, and SMOTE-Tomek. Random Forest (RF) classifier is chosen as the base learning algorithm for classification. The results from ensemble approach SRN-BRF are compared with other popular ensemble algorithms like RUSBoost, BRF, Over-Bagging, Over-Boost, SMOTE-Bagging, Balanced-Bagging, and Easy ensemble classifier. Five different performance metrics – accuracy, sensitivity, specificity, g-mean, and roc-auc score have been utilized for evaluation. Since classification accuracy can be quite misleading in the case of imbalanced datasets, a comparison among the approaches has been carried out based on the g-mean score and roc-auc score. It has been observed that the proposed algorithms outperformed all the other approaches in almost all datasets, in several cases by a large margin. The model's consistency in providing optimal results and its exceptional performance on highly skewed datasets substantiates the robustness of the proposed approaches and their supremacy over other sampling techniques.

**Table 1:** Overview of sampling techniques

| Algorithm name | Type | Short description of the methodology |
|---|---|---|
| SMOTE [1] | Oversampling | Uses interpolation to synthesize new samples from closely placed minority class samples |
| Borderline-SMOTE [4] | Oversampling | Extension of SMOTE focusing on borderline samples as they are more likely to be misclassified. |
| Safe-Level-SMOTE [6] | Oversampling | Extension of SMOTE. Assigns a safety level to the minority class samples based on its nearest neighbors. New samples are created only on the safe positions. |
| DBSMOTE [5] | Oversampling | Extension of SMOTE which utilizes DBSCAN clustering algorithm to improve minority class samples detection rate. |
| CURE-SMOTE [16] | Oversampling | Combining Clustering Using Representatives (CURE) with SMOTE algorithm |
| MWSMOTE [17] | Oversampling | Focusing on the shortcomings of the SMOTE algorithm, this majority weighted minority oversampling technique attempts to improve the sample selection and generation scheme. |
| ADASYN [3] | Oversampling | Adaptively generate samples based on their distribution. |
| IRUS [18] | Undersampling | This inverse random undersampling technique utilizes bagging method where the imbalance between the classes is reversed in different subsets of the data. |
| CBEUS [19] | Undersampling | This algorithm integrates clustering method (k-means clustering) with genetic algorithm to remove majority class samples that are far away from the centroid of each group. |
| ACOSampling [20] | Undersampling | This is a heuristic undersampling method based on ant-colony optimization algorithm. |
| OSS [9] | Hybrid Undersampling | This method combines Tomek-link based undersampling technique with CNN. |
| SMOTE-Tomek [12] | Hybrid | Combines SMOTE-based oversampling with Tomek-link based undersampling. |
| Random Balance [21] | Hybrid ensemble | Uses the boosting framework where undersampling and oversampling techniques are combined with random sampling ratios on different subsets. |
| RUSBoost [22] | Ensemble | Uses the boosting framework with randomly undersampled subsets. |
| Balanced Bagging [23] | Ensemble | Uses the bagging framework with randomly undersampled subsets. |
| BRF [24] | Ensemble | Uses the bagging framework with randomly undersampled subsets. Uses the RF architecture for training and classification. |
| SMOTE-Bagging [25] | Ensemble | Generates minority class instances using the SMOTE algorithm to balance the bootstrap samples. Uses bagging framework. |
| Easy Ensemble [26] | Hybrid Ensemble | Uses both the bagging and boosting framework. It is a bagging ensemble of AdaBoost classifiers each trained balanced data produced by the RUS approach. |
| Hard Ensemble [27] | Hybrid Ensemble | It is similar to easy ensemble but both oversampling and undersampling is used to balance the data. |

# 2. Methodology

## 2.1. Proposed framework

Sampling is a standard data preprocessing step when dealing with imbalanced data. However, this needs to be done with great care since misusing this process leads to data leakage, which generates over-optimistic results [28]. This occurs when the entire dataset is resampled before splitting it into training and testing sets. If the splitting is done after sampling, the original distribution of the data is lost and the class imbalance is no longer present in the testing set, leading to biased results. Also, some of the synthetic samples would be present in the test set, which is a typical data leakage issue. To avoid these pitfalls, sampling was performed after partitioning the data into training and test sets. To ensure better generalization, a 10-fold stratified cross-validation scheme was adopted for partitioning. The stratified strategy splits the data into training and test folds while maintaining the class imbalance ratio of the original data. This ensures that the original distribution of the data remains unchanged during the testing phase. Sampling techniques are then applied to the training folds, the RF classifier is trained on the sampled data, and its performance is measured on the testing folds. The average of the performances on the 10 testing folds is considered as the result.

A novel sampling algorithm (SMOTE-RUS-NC) is proposed in this study. The intuition behind the proposed algorithm is that sampling techniques like undersampling or oversampling do not always perform well on imbalanced data when used unilaterally. Especially when the imbalance ratio is large, these techniques tend to perform very poorly. A proper hybridization between oversampling and undersampling can improve the overall performance. This can be achieved using the proposed SMOTE-RUS-NC algorithm.

The sampling technique is further incorporated into the ensemble learning framework. The bagging approach uses bootstrap sampling to reduce variance. In our case, each of the bootstrap samples generated is balanced using the SMOTE-RUS-NC algorithm. We used a random subset of features while building each tree for diversity. This random forest architecture is able to provide better generalization with reduced variance. Instead of using only RUS to balance the bootstrap samples like the BRF classifier, a hybridization of three different sampling techniques is utilized here for balancing. This avails the benefits of the proposed SMOTE-RUS-NC architecture over the RUS approach and provides a more robust classification algorithm.

The outline of the proposed framework is illustrated in Figure 1.

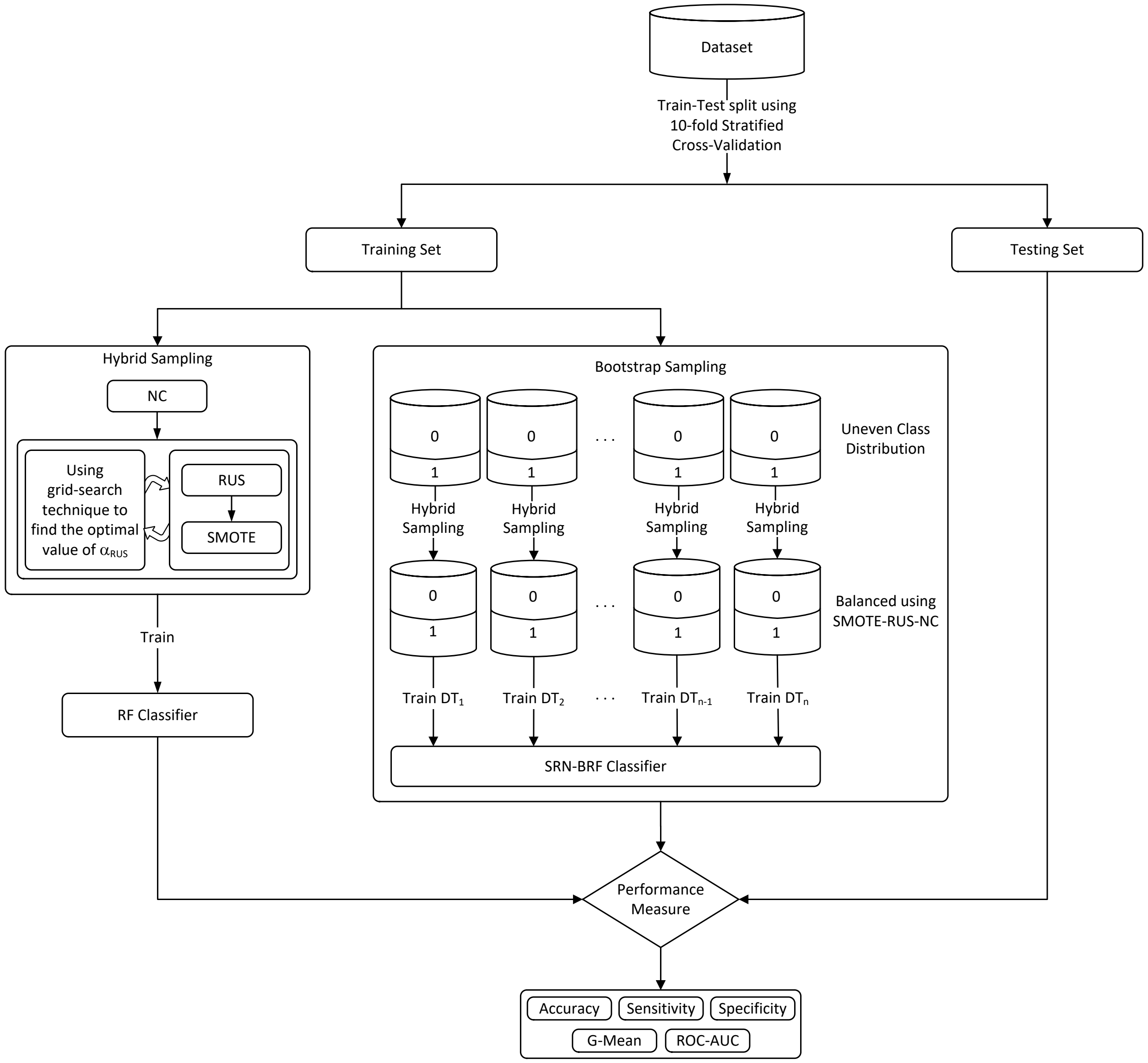

Figure 1. Outline of the proposed framework

## 2.2. SMOTE-RUS-NC

NC is a modification of the original Edited Nearest Neighbor (ENN) rule [15]. It provides a more in-depth data cleaning. For each sample in the training set ($s_i$), its $k_{nc}$ – nearest neighbors are located. $k_{nc}$ value of 3 is generally used. Using these $k_{nc}$ nearest neighbors, the original sample $s_i$ is classified. Now, if the original sample belongs to the majority class and is misclassified by its nearest neighbors – then that sample is removed from the training set. And if the original sample belongs to the minority class and is misclassified by its nearest neighbors – then the nearest neighbors that belong to the majority class are removed from the training set. This way, the majority class samples that can be considered noisy are identified and eliminated. By applying NC first, the imbalance between the classes is decreased. However, the number of samples removed in this process is generally not enough to obtain balance in the dataset. Even after applying NC, the dataset can still remain quite imbalanced.

The process is followed by the RUS algorithm. RUS randomly removes the majority class samples from the training set. This increases the chance of information loss as potentially useful data can be discarded in this process. The sampling ratio ($\alpha_{rus}$) may be adjusted to reduce information loss. Here, $\alpha_{rus}$ is defined as the ratio between the number of samples in the minority class ($N_{min}$) and the number of samples in the majority class after resampling ($N_{rus_maj}$).

$$\alpha_{rus} = \frac{N_{min}}{N_{rus_maj}} \tag{1}$$

The value of $\alpha_{rus}$ should be less than or equal to one. When $\alpha_{rus} = 1$ ($N_{min} = N_{rmaj}$), the RUS algorithm balances the dataset by removing the necessary number of majority class samples, which can be substantial in the case of highly imbalanced datasets. For example, if the dataset contains only 100 minority class samples and 10000 majority class samples – the RUS algorithm would eliminate 9900 samples to balance the dataset. Removing that number of samples is undesirable. Therefore, $\alpha_{rus}$ value should be adjusted to control how many samples are to be removed.

These two undersampling processes are then followed by the oversampling approach – SMOTE. To generate samples using SMOTE, first, a minority class sample ($s_i$) is selected from the training set. This selection is performed randomly. Then its $k_{smote}$ – nearest neighbors are located. $k_{smote}$ value of 5 is generally used. Among these $k_{smote}$ neighbors, one neighbor ($s_j$) is randomly chosen and used to synthesize a new sample using interpolation. The difference between the feature vector of $s_i$ and $s_j$ is calculated and multiplied by a random number between 0 and 1. This leads to the selection of a random point along the line between the features, which is the new synthetic sample. The number of samples to be generated by this algorithm can be controlled by the $\alpha_{smote}$ parameter. $\alpha_{smote}$ is the ratio of the number of samples in the minority class after sampling $N_{smote_min}$ to the number of samples in the majority class $N_{maj}$.

$$\alpha_{smote} = \frac{N_{smote_min}}{N_{maj}} \tag{2}$$

To avoid removing too many samples from the data, we do not balance the dataset by the combined undersampling techniques. The balance is obtained by generating new minority class samples using SMOTE. The number of samples to be synthesized is controlled by the $\alpha_{rus}$ parameter. Choosing a small value for $\alpha_{rus}$ (0.2 for instance), would remove a small number of samples from the majority class. This will in turn require a considerable number of samples to be generated by the SMOTE algorithm. On the contrary, setting a large value for $\alpha_{rus}$ (0.8 for instance), would remove too many samples from the majority class. Somewhere in the middle is the desirable point. From the experimentation, it has been observed that a value of 0.5 (set as default) would usually provide sub-optimal performance from the classifier. The grid-search technique can also be adopted to obtain the optimal value for $\alpha_{rus}$.

One thing to consider here is that although the NC algorithm is utilized in this study to bring down the class imbalance, it is not obligatory. Any other heuristic undersampling algorithm can be used here in its place. The idea is to lower the class imbalance first using a heuristic approach so that the number of samples to be removed by the RUS algorithm is minimized. Once the imbalance ratio is reduced, oversampling and undersampling are used in conjunction to obtain an optimal balance.

The proposed SMOTE-RUS-NC algorithm is illustrated in Figure 2 and summarized in the following section.

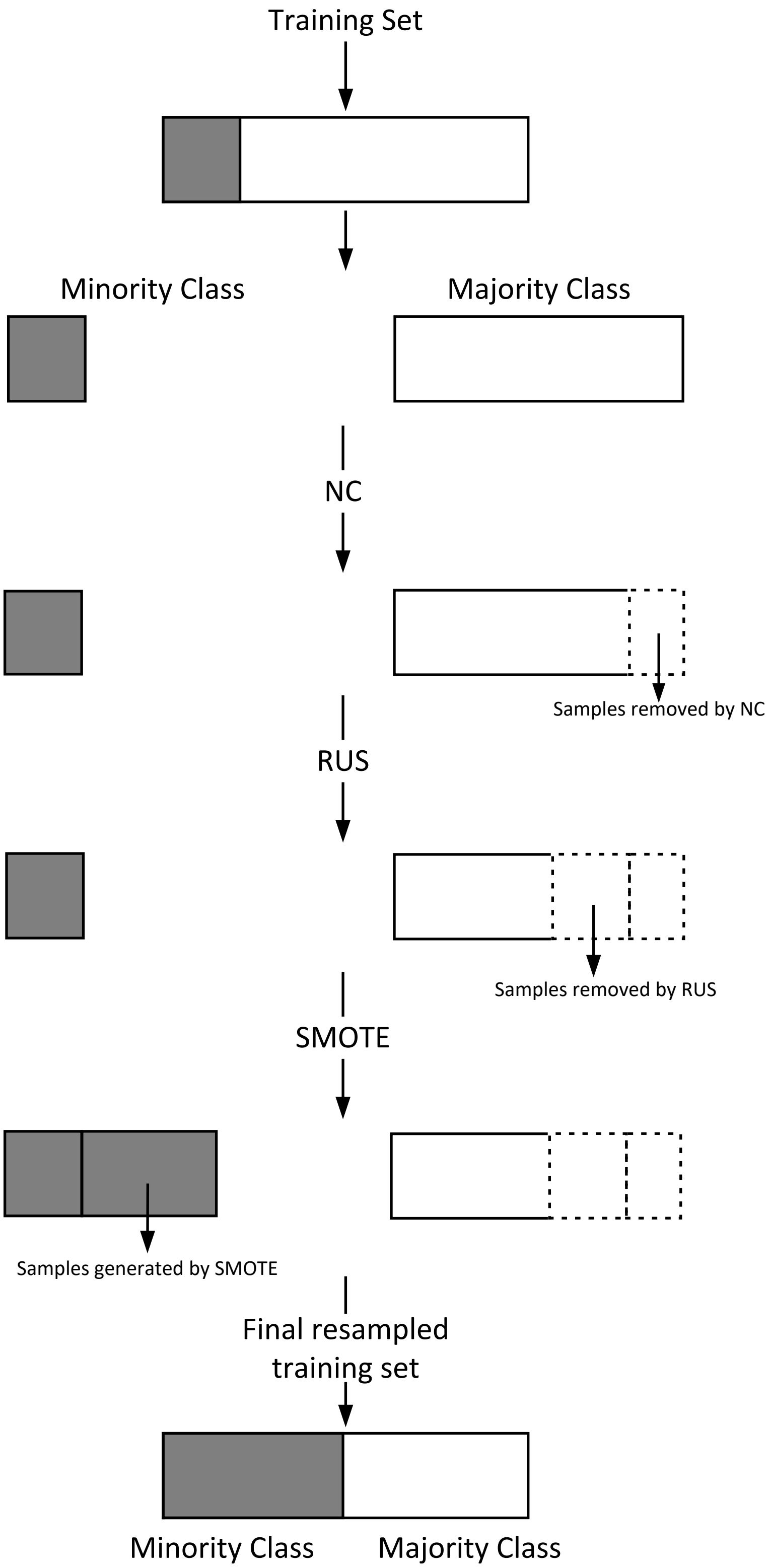

Figure 2. SMOTE-RUS-NC algorithm

## *Algorithm: SMOTE-RUS-NC*

Input:

- data (training set)
- $\alpha_{rus}$ – sampling ratio (default value is set to 0.5)
- $k_{nc}$ - the size of the neighborhood to consider while using NC (default value is set to 3)
- $k_{smote}$ - number of nearest neighbors used to generate synthetic samples using SMOTE (default value is set to 5)

Output:

- Resampled data

Function:

i. Calculate the number of samples in the minority class ($N_{min}$) and majority class ($N_{maj}$).

ii. Calculate the imbalance ratio, $\rho = \frac{N_{maj}}{N_{min}}$.

iii. Apply the NC algorithm to eliminate the noisy majority class samples from the training set. Unless $k_{nc}$ value is specified, it will use a 3-NN approach by default. The process can be parallelized for fast implementation.

iv. Calculate the reduced number of samples in the majority class ($N_{nc_maj}$). Then find the imbalance ratio, $\rho_{nc} = \frac{N_{nc_maj}}{N_{min}}$.

v. Select a set of values for $\alpha_{rus}$ such that $\frac{N_{min}}{\alpha_{rus}} < N_{nc_maj}$. For instance, $\alpha_{rus}$= [0.3, 0.4, 0.5, 0.6] – provided all these values satisfy the given criteria. The difference between the sampling points can be reduced to 0.05. This will obviously take more time; but might provide better results. It should be decided depending on the sample (dataset) size. This $\alpha_{rus}$ is an important hyperparameter of the proposed algorithm as it can affect the overall performance. So, tuning it can be proven beneficial.

vi. *for sampling_ratio in* $\alpha_{rus}$ *:*

   a. *apply the RUS algorithm to eliminate majority class samples such that –*

$$N_{rus_maj} = \frac{N_{min}}{sampling_ratio} \tag{3}$$

$$here, N_{rus_maj}\ is\ the\ number\ of\ samples\ after\ undersampling.$$

b. *Generate the required number of minority class samples using the SMOTE algorithm such that –*

$$N_{smote_min} = N_{rus_maj} \tag{4}$$

end for

end function

Although equation-4 has been maintained during experimentation on the datasets, this is not compulsory. $\mathrm{N_{smote_min}}$ value can be kept lower or higher than the $\mathrm{N_{rus_maj}}$ value. This can be controlled using the $\alpha_{smote}$ parameter. However, the effect of this parameter on the performance was found to be limited. Therefore, to maintain simplicity, only one hyperparameter was tuned ($\alpha_{rus}$) and the value of $\alpha_{smote}$ was kept as 1.

### 2.3. SRN-BRF

The original BRF classifier uses bootstrap aggregating or bagging, where each bootstrap subset of the data is balanced using the RUS algorithm. In our proposed methodology, instead of using RUS to balance the class distribution, we use the above-described SMOTE-RUS-NC algorithm. Each bootstrap sample is balanced using a three-step sampling process. Consequently, the number of samples removed by the RUS approach is minimized, reducing loss of information. The balanced bootstrap sample is then used to train a DT classifier. A total of 100 DTs are used to form the ensemble, each trained on a different balanced bootstrap subset of the data. The RF architecture is utilized as the ensemble mechanism. By balancing the class distribution this way, then merging it with the ensemble framework reduces the variance while also improving the performance. The proposed architecture of the ensemble algorithm SRN-BRF has been illustrated in Figure 1.

## 3. Experimental Setup

### 3.1. Data collection

To properly evaluate the performance of the proposed method, rigorous experimentation has been carried out on 26 benchmark imbalanced datasets. These datasets are collected from the KEEL-dataset repository [29] and the UCI repository [30]. All the datasets that are selected for experimentation are imbalanced with a minimum imbalance ratio of 1.8. To evaluate the efficacy of the proposed method in diverse imbalance setups, datasets with varying degrees of imbalance are selected. In several imbalanced datasets on the KEEL repository, the RF classifier trained on unsampled data provided exceptionally good performance (g-mean score above 99%). Since the classifier is already performing well on unsampled data, there is no additional need for sampling. So, comparison among sampling techniques on those datasets is groundless. Therefore, those datasets were not included in the experimentation.

Characteristics of the datasets utilized in this study are provided in Table 2.

**Table 2:** Characteristics of the datasets

| Dataset Name | No. of Samples | No. of Features | Imbalance Ratio |
|---|---|---|---|
| wisconsin | 683 | 9 | 1.86 |
| yeast1 | 1484 | 8 | 2.46 |
| vehicle1 | 846 | 18 | 2.9 |
| ecoli2 | 336 | 7 | 5.46 |
| yeast3 | 1484 | 8 | 8.1 |
| ecoli3 | 336 | 7 | 8.6 |
| page-blocks0 | 5472 | 10 | 8.79 |
| vowel0 | 988 | 13 | 9.98 |
| glass2 | 214 | 9 | 11.59 |
| glass4 | 214 | 9 | 15.47 |
| ecoli4 | 336 | 7 | 15.8 |
| abalone9-18 | 731 | 8 | 16.4 |
| flare-F | 1066 | 11 | 23.79 |
| yeast4 | 1484 | 8 | 28.1 |
| winequality-red-4 | 1599 | 11 | 29.17 |
| yeast-1-2-8-9_vs_7 | 947 | 8 | 30.57 |
| yeast5 | 1484 | 8 | 32.73 |
| ecoli-0-1-3-7_vs_2-6 | 281 | 7 | 39.14 |
| yeast6 | 1484 | 8 | 41.4 |
| abalone-19_vs_10-11-12-13 | 1622 | 8 | 49.69 |
| winequality-white-3-9_vs_5 | 1482 | 11 | 58.28 |
| poker-8-9_vs_6 | 1485 | 10 | 58.4 |
| winequality-red-3_vs_5 | 691 | 11 | 68.1 |
| abalone-20_vs_8-9-10 | 1916 | 8 | 72.69 |
| poker-8_vs_6 | 1485 | 10 | 85.88 |
| abalone19 | 4174 | 8 | 129.44 |

## 3.2. Performance Metrics

Five different metrics are utilized to evaluate the performance – accuracy, sensitivity, specificity, g-mean, and ROC-AUC score. Classification accuracy is the ratio between the total number of correct predictions and the total number of samples. This metric does not take into consideration the disparity between the classes. So, a classifier performing very well on the majority class prediction but performing poorly on minority class prediction would still attain a high accuracy value. Therefore, accuracy is not the proper metric to evaluate the performance in imbalance scenarios. Sensitivity (or recall) is a measure of how many positive samples are correctly classified, while specificity is a measure of how many negative samples are correctly classified. These two metrics represent how the algorithm is performing on individual classes. The geometric mean (g-mean) of these two metrics combines them into /a single value that represents the algorithm's overall performance. This metric is quite useful in evaluating the performance on imbalanced datasets, as bias towards any particular class results in a poor g-mean score. Another popular metric that was used is the ROC-AUC score (area under the ROC curve). ROC curve is a representation of true positive rate (sensitivity) vs. false positive rate. Using a scalar value of the area under this curve, performance comparison between different approaches can be carried out efficiently.

## 3.3. Classification algorithm

The Random Forest classifier is used in this study as the classification algorithm. This is a popular ensemble approach that often provides superior performance over other classification algorithms. It is an ensemble of decision trees, each of which is trained on bootstrapped data. Features used to build each tree are also randomized for better generalization. Decision trees are usually very prone to overfitting. They are models with medium bias and high variance. By aggregating multiple decision trees, better generalization can be achieved. In RF, randomness is inserted into the classifier by using bootstrapping and random feature subsets. A large forest with random trees overcomes the overfitting problem and provides a robust classification method. The number of trees used in this study is 100. Scikit-learn library's 'RandomForestClassifier' class with default settings was used during the experiment.

# 4. Results and discussion

A 10-fold stratified cross-validation scheme was undertaken for partitioning the data into training and test sets. The average of the performance on 10 testing folds is considered as the result. The performance of the proposed sampling algorithm SMOTE-RUS-NC is compared with several other popular sampling techniques. This includes oversampling techniques (SMOTE and ADASYN), undersampling techniques (RUS, NC, and CNN), and hybrid techniques (SMOTE-ENN and SMOTE-Tomek). The same workflow, as well as the same train-test partition, was utilized for each of these techniques. The parameters of the RF classifier were also kept the same for all cases. The performance of the proposed ensemble algorithm SRN-BRF is compared with other popular ensemble techniques used in the imbalanced domain. This includes Over-Bagging, SMOTE-Bagging, Over-Boost, RUSBoost, Balanced Bagging (BB), BRF, and Easy Ensemble classifier.

## 4.1. Performance of the RF classifier on unsampled data

The performance measures obtained are reported in Tables 3 to 7. As can be observed from the results, the accuracy values obtained in different datasets using a standard RF classifier without any sampling are quite high. However, this high accuracy value is quite misleading, which is apparent from the difference in the sensitivity and specificity scores. For instance, in the yeast4 dataset, the sensitivity score obtained is only 14% while the specificity score is 99.65%. Although the accuracy is quite high at 96.7%, the classifier is clearly failing to distinguish the minority class samples from the majority class samples. The g-mean score provides a more literal picture of the high bias caused by the imbalance scenario. On unsampled data, the g-mean score obtained is comparatively low in most datasets. There is a major decline in performance with an increased imbalance ratio. In some datasets, the scores fall below 10%. In few of the datasets, the g-mean score was completely 0, indicating a complete bias towards the majority class.

## 4.2. Performance of the RF classifier after resampling the data with state-of-the-art sampling techniques

Using sampling techniques can improve the performance of the classifier. For some datasets, the improvement is remarkable. While on some other datasets, the improvement is small. For instance, on the abalone9-18 dataset, the g-mean score increased to 67.43% from 26.49% after SMOTE was applied. While on the glass2 dataset, the g-mean score increased to only 26.56% from 7.07% using SMOTE. On the abalone-19_vs_10-11-12-13 dataset, even after sampling, the g-mean score obtained is 0, indicating the incompetence of the SMOTE algorithm to counter bias. This indicates that the SMOTE algorithm is not always able to provide a satisfactory improvement in performance. Similar scenarios are observed with other sampling algorithms as well. On the winequality-red-3_vs_5 dataset, all the other sampling algorithms except CNN failed to counter the bias, providing a g-mean score of 0 (CNN attained 19.78% only). On the yeast-1-2-8-9_vs_7 dataset, applying SMOTE-ENN-based hybrid sampling improved the performance from 19.67% to 31.88% only. While on the same dataset, there is a drop in performance (17.13%) when NC-based undersampling was employed.

In summary, the sampling algorithms improve the performance of the classifier to different extents. However, with an increased imbalance ratio, the performance improvement is limited. In the worst-case scenarios, the performance might also deteriorate. In several cases, the sampling techniques completely failed to counter the bias, resulting in a g-mean score of 0. Moreover, the same sampling technique is not always able to offer good performance in all datasets and there is no definite way to know which one of these approaches would be able to provide the best performance.

**Table 3:** Performance comparison of the proposed sampling algorithm with other sampling techniques in terms of accuracy (in percentage)

| **Dataset** | **Imbalance Ratio** | **Unsampled** | **SMOTE** | **ADASYN** | **RUS** | **NC** | **CNN** | **SMOTEENN** | **SMOTE-TOMEK** | **SMOTE-RUS-NC** |
|---|---|---|---|---|---|---|---|---|---|---|
| wisconsin | 1.86 | 96.79 | 96.49 | 96.63 | 97.22 | 97.66 | 96.49 | 97.08 | 97.23 | 97.51 |
| yeast1 | 2.46 | 94.88 | 95.01 | 94.88 | 92.66 | 95.35 | 94.88 | 94.07 | 94.94 | 93.13 |
| vehicle1 | 2.90 | 78.48 | 78.37 | 77.66 | 76.12 | 76.24 | 77.30 | 73.40 | 78.25 | 77.31 |
| ecoli2 | 5.46 | 90.37 | 91.27 | 89.20 | 87.33 | 91.57 | 90.36 | 89.16 | 90.06 | 89.46 |
| yeast3 | 8.10 | 94.67 | 95.28 | 94.81 | 91.78 | 95.01 | 95.21 | 94.20 | 95.14 | 93.12 |
| ecoli3 | 8.60 | 93.16 | 91.63 | 90.13 | 80.84 | 91.05 | 92.24 | 86.53 | 91.02 | 88.64 |
| page-blocks0 | 8.79 | 96.80 | 96.07 | 95.49 | 93.02 | 96.14 | 96.36 | 95.19 | 96.03 | 93.93 |
| vowel0 | 9.98 | 96.36 | 97.07 | 96.56 | 96.77 | 97.06 | 96.26 | 97.88 | 97.98 | 97.37 |
| glass2 | 11.59 | 92.51 | 86.82 | 87.29 | 60.71 | 90.65 | 87.27 | 80.30 | 85.45 | 70.95 |
| glass4 | 15.47 | 96.73 | 97.66 | 96.71 | 84.78 | 95.30 | 89.63 | 95.28 | 97.19 | 91.93 |
| ecoli4 | 15.80 | 98.21 | 97.59 | 97.29 | 91.85 | 97.60 | 92.19 | 96.68 | 97.90 | 92.14 |
| abalone9-18 | 16.40 | 94.80 | 91.66 | 90.98 | 72.76 | 94.39 | 93.44 | 87.14 | 91.93 | 81.53 |
| flare-F | 23.79 | 93.62 | 93.81 | 93.06 | 79.16 | 93.62 | 90.24 | 89.95 | 93.15 | 85.25 |
| yeast4 | 28.10 | 96.70 | 95.35 | 95.15 | 78.69 | 96.76 | 96.02 | 92.11 | 95.42 | 83.55 |
| winequality-red-4 | 29.17 | 96.62 | 92.49 | 92.74 | 69.28 | 96.50 | 92.81 | 89.61 | 92.87 | 76.17 |
| yeast-1-2-8-9_vs_7 | 30.57 | 96.83 | 95.02 | 94.81 | 70.39 | 96.09 | 96.72 | 91.00 | 94.71 | 82.64 |
| yeast5 | 32.73 | 98.25 | 98.11 | 98.11 | 93.59 | 98.45 | 98.58 | 98.11 | 98.11 | 96.63 |
| ecoli-0-1-3-7_vs_2-6 | 39.14 | 97.96 | 97.45 | 97.45 | 71.94 | 97.96 | 97.96 | 95.92 | 97.45 | 83.67 |
| yeast6 | 41.40 | 98.11 | 97.44 | 96.97 | 89.35 | 98.04 | 98.18 | 96.16 | 97.44 | 88.40 |
| abalone-19_vs_10-11-12-13 | 49.69 | 98.09 | 98.09 | 95.31 | 95.31 | 98.09 | 97.59 | 98.09 | 95.50 | 82.79 |
| winequality-white-3-9_vs_5 | 58.28 | 98.25 | 98.80 | 95.88 | 96.09 | 98.04 | 97.16 | 98.04 | 96.63 | 74.95 |
| poker-8-9_vs_6 | 58.40 | 98.38 | 98.38 | 99.39 | 99.33 | 98.38 | 98.65 | 98.38 | 99.33 | 97.85 |
| winequality-red-3_vs_5 | 68.10 | 98.26 | 98.26 | 95.51 | 96.09 | 98.26 | 96.38 | 98.26 | 96.38 | 81.30 |
| abalone-20_vs_8-9-10 | 72.69 | 98.69 | 98.69 | 97.70 | 97.75 | 98.69 | 97.96 | 98.69 | 97.60 | 90.27 |
| poker-8_vs_6 | 85.88 | 98.85 | 98.85 | 99.53 | 99.53 | 98.92 | 98.92 | 98.92 | 99.53 | 92.75 |
| abalone19 | 129.44 | 99.23 | 99.23 | 97.24 | 97.05 | 99.23 | 99.19 | 99.23 | 97.29 | 77.79 |

**Table 4:** Performance comparison of the proposed sampling algorithm with other sampling techniques in terms of sensitivity (in percentage)

| Dataset | Imbalance Ratio | Unsampled | SMOTE | ADASYN | RUS | NC | CNN | SMOTEENN | SMOTE-TOMEK | SMOTE-RUS-NC |
|---|---|---|---|---|---|---|---|---|---|---|
| wisconsin | 1.86 | 95.40 | 96.23 | 97.90 | 97.90 | 98.73 | 97.48 | 97.90 | 97.48 | 99.57 |
| yeast1 | 2.46 | 69.85 | 80.88 | 85.18 | 92.57 | 83.38 | 81.47 | 88.27 | 81.51 | 93.79 |
| vehicle1 | 2.90 | 48.35 | 67.73 | 66.28 | 82.42 | 78.83 | 70.00 | 88.03 | 64.89 | 88.96 |
| ecoli2 | 5.46 | 75.00 | 83.00 | 86.67 | 88.67 | 85.00 | 88.33 | 88.67 | 84.67 | 90.33 |
| yeast3 | 8.10 | 69.78 | 83.27 | 85.15 | 90.66 | 82.68 | 83.31 | 88.86 | 82.65 | 93.82 |
| ecoli3 | 8.60 | 52.50 | 74.17 | 65.83 | 94.17 | 60.00 | 62.50 | 89.17 | 70.83 | 91.67 |
| page-blocks0 | 8.79 | 83.20 | 88.56 | 89.10 | 93.92 | 88.20 | 86.78 | 91.78 | 88.03 | 95.38 |
| vowel0 | 9.98 | 67.78 | 80.00 | 72.22 | 90.00 | 73.33 | 94.44 | 86.67 | 86.67 | 97.78 |
| glass2 | 11.59 | 5.00 | 20.00 | 30.00 | 65.00 | 10.00 | 30.00 | 30.00 | 20.00 | 85.00 |
| glass4 | 15.47 | 65.00 | 80.00 | 80.00 | 95.00 | 65.00 | 80.00 | 90.00 | 90.00 | 95.00 |
| ecoli4 | 15.80 | 70.00 | 80.00 | 80.00 | 90.00 | 70.00 | 90.00 | 85.00 | 85.00 | 100 |
| abalone9-18 | 16.40 | 14.50 | 49.50 | 45.50 | 73.00 | 21.50 | 36.00 | 51.00 | 43.00 | 70.50 |
| flare-F | 23.79 | 96.97 | 95.79 | 95.50 | 78.77 | 96.97 | 91.49 | 91.29 | 95.30 | 85.32 |
| yeast4 | 28.10 | 14.00 | 47.00 | 43.33 | 90.00 | 26.00 | 29.67 | 58.67 | 43.33 | 84.00 |
| winequality-red-4 | 29.17 | 99.94 | 95.21 | 95.28 | 69.78 | 99.81 | 95.48 | 91.91 | 95.67 | 76.72 |
| yeast-1-2-8-9_vs_7 | 30.57 | 13.33 | 20.00 | 20.00 | 76.67 | 10.00 | 26.67 | 23.33 | 13.33 | 66.67 |
| yeast5 | 32.73 | 57.50 | 84.00 | 86.50 | 98.00 | 78.50 | 84.50 | 91.00 | 84.50 | 98.00 |
| ecoli-0-1-3-7_vs_2-6 | 39.14 | 42.86 | 57.14 | 57.14 | 85.71 | 42.86 | 57.14 | 57.14 | 57.14 | 85.71 |
| yeast6 | 41.40 | 35.00 | 58.33 | 58.33 | 82.50 | 40.83 | 58.33 | 65.83 | 55.83 | 87.50 |
| abalone-19_vs_10-11-12-13 | 49.69 | 0.00 | 0.00 | 16.67 | 16.67 | 0.00 | 0.00 | 0.00 | 20.00 | 58.33 |
| winequality-white-3-9_vs_5 | 58.28 | 3.33 | 0.00 | 13.33 | 13.33 | 10.00 | 8.33 | 10.00 | 13.33 | 76.67 |
| poker-8-9_vs_6 | 58.40 | 3.33 | 3.33 | 63.33 | 58.33 | 3.33 | 21.67 | 3.33 | 58.33 | 86.67 |
| winequality-red-3_vs_5 | 68.10 | 0.00 | 0.00 | 0.00 | 0.00 | 0.00 | 20.00 | 0.00 | 0.00 | 70.00 |
| abalone-20_vs_8-9-10 | 72.69 | 8.33 | 8.33 | 46.67 | 51.67 | 16.67 | 11.67 | 16.67 | 48.33 | 81.67 |
| poker-8_vs_6 | 85.88 | 0.00 | 0.00 | 55.00 | 55.00 | 5.00 | 5.00 | 5.00 | 55.00 | 70.00 |
| abalone19 | 129.44 | 0.00 | 0.00 | 15.83 | 12.50 | 0.00 | 0.00 | 0.00 | 12.50 | 71.67 |

**Table 5:** Performance comparison of the proposed sampling algorithm with other sampling techniques in terms of specificity (in percentage)

| Dataset | Imbalance Ratio | Unsampled | SMOTE | ADASYN | RUS | NC | CNN | SMOTEENN | SMOTE-TOMEK | SMOTE-RUS-NC |
|---|---|---|---|---|---|---|---|---|---|---|
| wisconsin | 1.86 | 97.54 | 96.63 | 95.95 | 96.86 | 97.09 | 95.96 | 96.64 | 97.09 | 96.41 |
| yeast1 | 2.46 | 97.96 | 96.74 | 96.06 | 92.66 | 96.82 | 96.52 | 94.78 | 96.59 | 93.03 |
| vehicle1 | 2.90 | 88.87 | 46.50 | 35.50 | 89.00 | 15.00 | 60.50 | 59.50 | 42.50 | 73.29 |
| ecoli2 | 5.46 | 93.21 | 92.87 | 89.74 | 87.16 | 92.86 | 90.73 | 89.32 | 91.11 | 89.32 |
| yeast3 | 8.10 | 97.73 | 96.74 | 95.98 | 91.89 | 96.52 | 96.67 | 94.85 | 96.67 | 93.03 |
| ecoli3 | 8.60 | 98.00 | 93.67 | 93.00 | 79.33 | 94.67 | 95.67 | 86.33 | 93.33 | 88.33 |
| page-blocks0 | 8.79 | 98.35 | 96.93 | 96.22 | 92.92 | 97.05 | 97.46 | 95.58 | 96.95 | 93.77 |
| vowel0 | 9.98 | 99.22 | 98.78 | 99.00 | 97.44 | 99.44 | 96.44 | 99.00 | 99.11 | 97.33 |
| glass2 | 11.59 | 100 | 92.21 | 92.21 | 60.29 | 97.45 | 92.29 | 84.13 | 90.74 | 69.97 |
| glass4 | 15.47 | 99.00 | 99.00 | 98.00 | 84.50 | 97.50 | 90.50 | 96.00 | 98.00 | 92.00 |
| ecoli4 | 15.80 | 100 | 98.71 | 98.39 | 91.96 | 99.35 | 92.32 | 97.42 | 98.71 | 91.63 |
| abalone9-18 | 16.40 | 99.71 | 94.20 | 93.76 | 72.71 | 98.84 | 96.95 | 89.25 | 94.92 | 82.14 |
| flare-F | 23.79 | 15.00 | 46.50 | 35.50 | 89.00 | 15.00 | 60.50 | 59.50 | 42.50 | 84.00 |
| yeast4 | 28.10 | 99.65 | 97.07 | 97.00 | 78.29 | 99.30 | 98.39 | 93.30 | 97.28 | 83.52 |
| winequality-red-4 | 29.17 | 0.00 | 13.33 | 18.67 | 54.67 | 0.00 | 15.00 | 22.67 | 11.33 | 60.67 |
| yeast-1-2-8-9_vs_7 | 30.57 | 99.56 | 97.48 | 97.26 | 70.18 | 98.90 | 99.02 | 93.21 | 97.37 | 83.16 |
| yeast5 | 32.73 | 99.51 | 98.54 | 98.47 | 93.46 | 99.10 | 99.03 | 98.33 | 98.54 | 96.59 |
| ecoli-0-1-3-7_vs_2-6 | 39.14 | 100 | 98.94 | 98.94 | 71.43 | 100 | 99.47 | 97.35 | 98.94 | 83.60 |
| yeast6 | 41.40 | 99.65 | 98.41 | 97.93 | 89.50 | 99.45 | 99.17 | 96.89 | 98.48 | 88.39 |
| abalone-19_vs_10-11-12-13 | 49.69 | 100 | 100 | 96.86 | 96.86 | 100 | 99.50 | 100 | 96.98 | 83.27 |
| winequality-white-3-9_vs_5 | 58.28 | 99.86 | 100 | 97.32 | 97.53 | 99.52 | 98.69 | 99.52 | 98.08 | 75.01 |
| poker-8-9_vs_6 | 58.40 | 100 | 100 | 100 | 100 | 100 | 100 | 100 | 100 | 98.08 |
| winequality-red-3_vs_5 | 68.10 | 99.71 | 99.71 | 96.91 | 97.50 | 99.71 | 97.50 | 99.71 | 97.79 | 81.47 |
| abalone-20_vs_8-9-10 | 72.69 | 99.95 | 99.95 | 98.41 | 98.41 | 99.84 | 99.15 | 99.84 | 98.31 | 90.41 |
| poker-8_vs_6 | 85.88 | 100 | 100 | 100 | 100 | 100 | 100 | 100 | 100 | 93.01 |
| abalone19 | 129.44 | 100 | 100 | 97.88 | 97.71 | 100 | 99.95 | 100 | 97.95 | 77.83 |

**Table 6:** Performance comparison of the proposed sampling algorithm with other sampling techniques in terms of G-Mean (in percentage)

| **Dataset** | **Imbalance Ratio** | **Unsampled** | **SMOTE** | **ADASYN** | **RUS** | **NC** | **CNN** | **SMOTEENN** | **SMOTE-TOMEK** | **SMOTE-RUS-NC** |
|---|---|---|---|---|---|---|---|---|---|---|
| wisconsin | 1.86 | 96.41 | 96.39 | 96.90 | 97.36 | 97.89 | 96.68 | 97.25 | 97.25 | **97.97** |
| yeast1 | 2.46 | 82.37 | 88.32 | 90.38 | 92.57 | 89.77 | 88.54 | 91.43 | 88.62 | **93.34** |
| vehicle1 | 2.90 | 64.99 | 74.41 | 73.35 | 78.00 | 76.98 | 74.65 | 77.45 | 73.10 | **80.60** |
| ecoli2 | 5.46 | 82.54 | 86.37 | 87.01 | 85.37 | 87.68 | 88.37 | 86.92 | 86.25 | **88.39** |
| yeast3 | 8.10 | 82.26 | 89.65 | 90.34 | 91.20 | 89.21 | 89.64 | 91.76 | 89.25 | **93.36** |
| ecoli3 | 8.60 | 70.45 | 82.63 | 77.15 | 86.26 | 74.33 | 76.81 | 87.39 | 80.78 | **89.70** |
| page-blocks0 | 8.79 | 89.96 | 92.38 | 92.33 | 93.20 | 92.12 | 91.58 | 93.43 | 92.09 | **94.45** |
| vowel0 | 9.98 | 75.36 | 87.44 | 81.74 | 93.02 | 79.49 | 94.86 | 91.86 | 92.02 | **97.47** |
| glass2 | 11.59 | 7.07 | 26.56 | 36.56 | 57.40 | 13.77 | 37.30 | 34.11 | 26.17 | **75.81** |
| glass4 | 15.47 | 71.21 | 83.64 | 83.13 | 86.86 | 70.70 | 82.06 | 92.09 | **93.13** | 92.86 |
| ecoli4 | 15.80 | 82.43 | 87.63 | 87.46 | 89.71 | 82.10 | 90.17 | 89.88 | 90.55 | **95.38** |
| abalone9-18 | 16.40 | 26.49 | 67.43 | 65.01 | 71.94 | 37.85 | 58.17 | 66.27 | 63.38 | **75.41** |
| flare-F | 23.79 | 19.25 | 64.22 | 53.04 | 83.24 | 19.25 | 72.30 | 71.40 | 61.42 | **84.25** |
| yeast4 | 28.10 | 28.69 | 66.17 | 63.48 | **83.15** | 47.10 | 47.21 | 72.96 | 63.56 | 83.12 |
| winequality-red-4 | 29.17 | 0.00 | 22.00 | 32.00 | 59.28 | 0.00 | 31.39 | 36.55 | 20.25 | **65.71** |
| yeast-1-2-8-9_vs_7 | 30.57 | 19.67 | 30.76 | 30.74 | 72.16 | 17.13 | 42.58 | 31.88 | 22.61 | **73.00** |
| yeast5 | 32.73 | 72.70 | 90.15 | 91.69 | 95.63 | 87.40 | 91.16 | 94.26 | 90.68 | **97.24** |
| ecoli-0-1-3-7_vs_2-6 | 39.14 | 42.86 | 56.61 | 56.61 | 68.52 | 42.86 | 56.88 | 55.79 | 56.61 | **76.75** |
| yeast6 | 41.40 | 51.90 | 70.36 | 70.17 | 85.50 | 55.84 | 74.35 | 78.14 | 68.31 | **87.54** |
| abalone-19_vs_10-11-12-13 | 49.69 | 0.00 | 0.00 | 24.95 | 24.96 | 0.00 | 0.00 | 0.00 | 27.33 | **68.65** |
| winequality-white-3-9_vs_5 | 58.28 | 5.77 | 0.00 | 19.55 | 19.70 | 17.26 | 12.64 | 17.26 | 19.70 | **74.08** |
| poker-8-9_vs_6 | 58.40 | 5.77 | 5.77 | 73.85 | 70.93 | 5.77 | 28.62 | 5.77 | 70.93 | **91.19** |
| winequality-red-3_vs_5 | 68.10 | 0.00 | 0.00 | 0.00 | 0.00 | 0.00 | 19.78 | 0.00 | 0.00 | **62.90** |
| abalone-20_vs_8-9-10 | 72.69 | 12.84 | 12.84 | 59.01 | 66.05 | 25.67 | 18.60 | 25.67 | 60.24 | **84.62** |
| poker-8_vs_6 | 85.88 | 0.00 | 0.00 | 65.36 | 65.36 | 7.07 | 7.07 | 7.07 | 65.36 | **75.32** |
| abalone19 | 129.44 | 0.00 | 0.00 | 27.74 | 22.01 | 0.00 | 0.00 | 0.00 | 22.04 | **73.39** |

**Table 7:** Performance comparison of the proposed sampling algorithm with other sampling techniques in terms of ROC-AUC (in percentage)

| **Dataset** | **Imbalance Ratio** | **Unsampled** | **SMOTE** | **ADASYN** | **RUS** | **NC** | **CNN** | **SMOTEENN** | **SMOTE-TOMEK** | **SMOTE-RUS-NC** |
|---|---|---|---|---|---|---|---|---|---|---|
| wisconsin | 1.86 | 96.47 | 96.43 | 96.93 | 97.38 | 97.91 | 96.72 | 97.27 | 97.28 | 97.99 |
| yeast1 | 2.46 | 83.90 | 88.81 | 90.62 | 92.61 | 90.10 | 88.99 | 91.52 | 89.05 | 93.41 |
| vehicle1 | 2.90 | 68.61 | 74.88 | 73.92 | 78.18 | 77.09 | 74.90 | 78.19 | 73.86 | 81.12 |
| ecoli2 | 5.46 | 84.11 | 87.93 | 88.20 | 87.91 | 88.93 | 89.53 | 88.99 | 87.89 | 89.83 |
| yeast3 | 8.10 | 83.75 | 90.01 | 90.57 | 91.28 | 89.60 | 89.99 | 91.85 | 89.66 | 93.43 |
| ecoli3 | 8.60 | 75.25 | 83.92 | 79.42 | 86.75 | 77.33 | 79.08 | 87.75 | 82.08 | 90.00 |
| page-blocks0 | 8.79 | 90.78 | 92.74 | 92.66 | 93.42 | 92.63 | 92.12 | 93.68 | 92.49 | 94.57 |
| vowel0 | 9.98 | 83.50 | 89.39 | 85.61 | 93.72 | 86.39 | 95.44 | 92.83 | 92.89 | 97.56 |
| glass2 | 11.59 | 52.50 | 56.11 | 61.11 | 62.64 | 53.72 | 61.14 | 57.07 | 55.37 | 77.49 |
| glass4 | 15.47 | 82.00 | 89.50 | 89.00 | 89.75 | 81.25 | 85.25 | 93.00 | 94.00 | 93.50 |
| ecoli4 | 15.80 | 85.00 | 89.35 | 89.19 | 90.98 | 84.68 | 91.16 | 91.21 | 91.85 | 95.82 |
| abalone9-18 | 16.40 | 57.11 | 71.85 | 69.63 | 72.86 | 60.17 | 66.47 | 70.13 | 68.96 | 76.32 |
| flare-F | 23.79 | 55.99 | 71.15 | 65.50 | 83.89 | 55.99 | 76.00 | 75.39 | 68.90 | 84.66 |
| yeast4 | 28.10 | 56.83 | 72.03 | 70.17 | 84.14 | 62.65 | 64.03 | 75.98 | 70.31 | 83.76 |
| winequality-red-4 | 29.17 | 49.97 | 54.27 | 56.97 | 62.23 | 49.90 | 55.24 | 57.29 | 53.50 | 68.69 |
| yeast-1-2-8-9_vs_7 | 30.57 | 56.45 | 58.74 | 58.63 | 73.43 | 54.45 | 62.84 | 58.27 | 55.35 | 74.91 |
| yeast5 | 32.73 | 78.51 | 91.27 | 92.49 | 95.73 | 88.80 | 91.76 | 94.67 | 91.52 | 97.30 |
| ecoli-0-1-3-7_vs_2-6 | 39.14 | 71.43 | 78.04 | 78.04 | 78.57 | 71.43 | 78.31 | 77.25 | 78.04 | 84.66 |
| yeast6 | 41.40 | 67.33 | 78.37 | 78.13 | 86.00 | 70.14 | 78.75 | 81.36 | 77.16 | 87.95 |
| abalone-19_vs_10-11-12-13 | 49.69 | 50.00 | 50.00 | 56.76 | 56.76 | 50.00 | 49.75 | 50.00 | 58.49 | 70.81 |
| winequality-white-3-9_vs_5 | 58.28 | 51.60 | 50.00 | 55.33 | 55.43 | 54.76 | 53.51 | 54.76 | 55.71 | 75.84 |
| poker-8-9_vs_6 | 58.40 | 51.67 | 51.67 | 81.67 | 79.17 | 51.67 | 60.83 | 51.67 | 79.17 | 92.37 |
| winequality-red-3_vs_5 | 68.10 | 49.85 | 49.85 | 48.46 | 48.75 | 49.85 | 58.75 | 49.85 | 48.90 | 75.74 |
| abalone-20_vs_8-9-10 | 72.69 | 54.14 | 54.14 | 72.54 | 75.04 | 58.25 | 55.41 | 58.25 | 73.32 | 86.04 |
| poker-8_vs_6 | 85.88 | 50.00 | 50.00 | 77.50 | 77.50 | 52.50 | 52.50 | 52.50 | 77.50 | 81.51 |
| abalone19 | 129.44 | 50.00 | 50.00 | 56.85 | 55.10 | 50.00 | 49.98 | 50.00 | 55.22 | 74.75 |

## 4.3. Performance of the RF classifier after resampling the data with the proposed SMOTE-RUS-NC algorithm

To overcome the limitations of those sampling techniques, a hybrid sampling algorithm is introduced in this study. It first brings down the imbalance ratio using the NC algorithm, then combines random undersampling with SMOTE-based oversampling method. By finding a suitable operating point, the best possible performance can be achieved. As can be observed from the results, the proposed SMOTE-RUS-NC algorithm outperformed all the other sampling techniques in almost all datasets (24 out of 26). On the yeast-1-2-8-9_vs_7 dataset, the g-mean score obtained using this algorithm is 73%, which is significantly higher than other algorithms like SMOTE (30.76%), or SMOTE-ENN (31.88%), or SMOTE-Tomek (22.68%). On the glass2 dataset, the g-mean score obtained using this algorithm is 75.81%. The second highest score obtained in this dataset is 57.4% using the RUS algorithm, which is comparatively much lower. In a number of different datasets, the SMOTE-RUS-NC method performs much better than any other sampling strategy. Especially on the highly imbalanced datasets, it provided exceptionally better results than any other sampling techniques. On the abalone19 dataset (IR = 129.44), the obtained g-mean score is 73.39%, while the second-highest score was 27.74% with the ADASYN algorithm. On the winequality-red-3_vs_5 dataset (IR = 68.1) where most of the other sampling techniques provided a g-mean score of 0, the proposed approach provided a g-mean score of 62.9%.

On some datasets, however, the improvement in performance compared to other sampling algorithms is small. This is observed mostly in low-imbalance scenarios like in the ecoli2 dataset (IR = 5.46). The RF classifier on unsampled data provided a g-mean score of 82.54%, while the proposed algorithm provided a g-mean score of 88.39%. On a couple of other datasets like glass4, the highest g-mean score was attained using the SMOTE-Tomek algorithm (93.13%). On this dataset, the proposed algorithm provided a g-mean score of 92.86%, which is less than 1% lower than that. Similarly, on the yeast4 dataset, the RUS approach provided the highest g-mean score of 83.15%, while the g-mean score obtained by the proposed approach is almost the same (83.12%).

One thing to consider while using the SMOTE-RUS-NC algorithm is the randomness associated with the RUS method. Since the RUS algorithm randomly removes samples from the majority class (although a limited number of samples are removed in case of SMOTE-RUS-NC), it can affect the overall performance of the classifier. So, the results obtained may vary (increase or decrease) by a small percentage with different seeds. Using a set of values for the $\alpha_{rus}$ parameter (utilized in this study) can help in obtaining optimal result.

The consistency of the algorithm’s performance is quite noticeable. While other techniques failed to perform well in several datasets, the proposed algorithm provided optimal results in all cases. Particularly in the case of high imbalance scenarios, the proposed algorithm displayed unparalleled performance compared to other approaches. Based on the results obtained, it can be concluded that the proposed algorithm is quite effective when dealing with imbalanced data. Sampling data using the SMOTE-RUS-NC algorithm yielded better results than other state-of-the-art sampling techniques, indicating its supremacy over other methods.

The difference in performance (based on G-mean score) among these sampling techniques on some highly imbalanced datasets is illustrated in Figure 3.

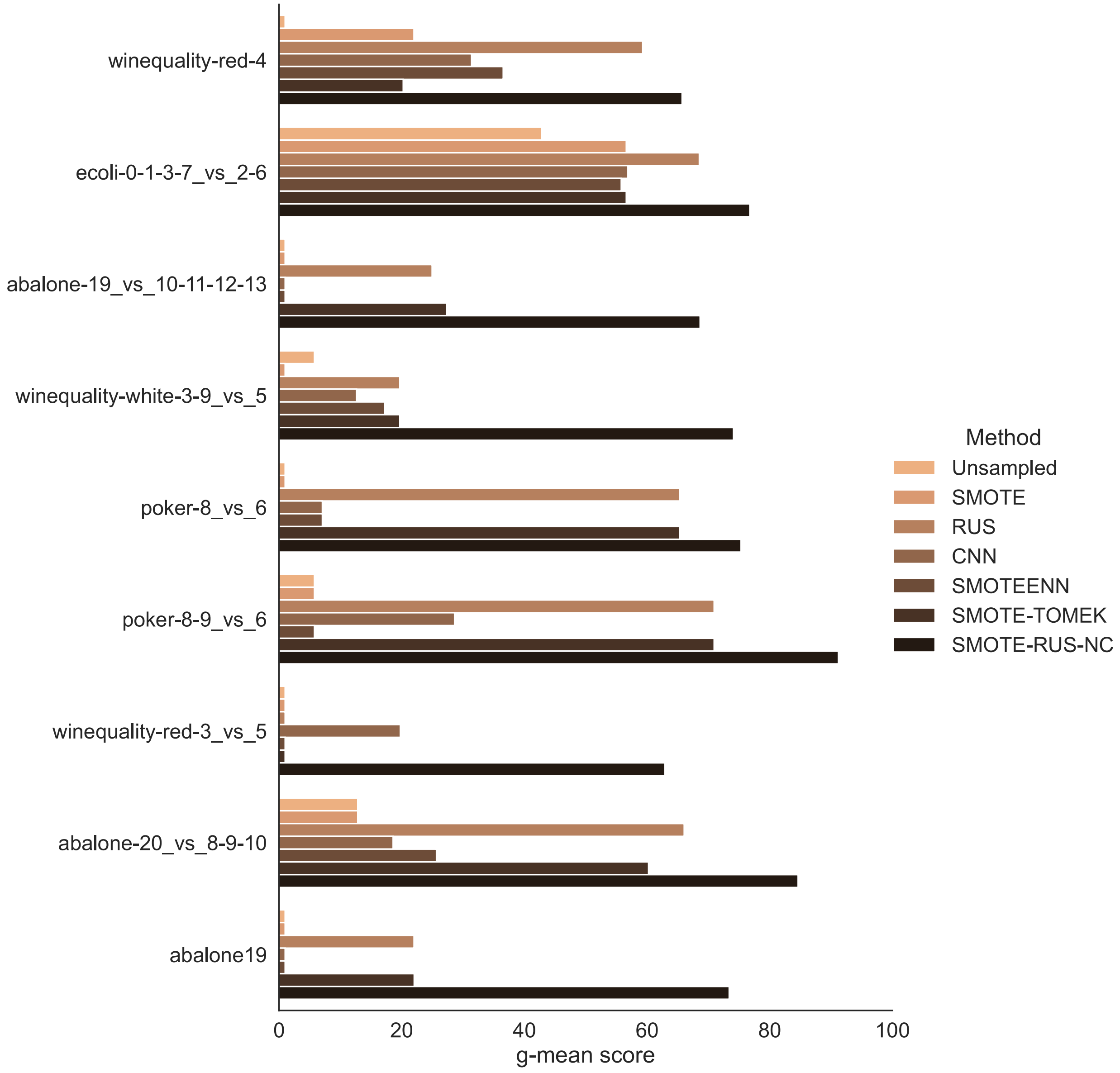

**Figure 3.** Performance comparison (in terms of G-mean score) of the proposed SMOTE-RUS-NC approach with other techniques (Unsampled, SMOTE, RUS, CNN, SMOTEENN, and SMOTE-TOMEK)

## 4.4. Performance of the proposed SRN-BRF classifier and its comparison with other ensemble algorithms

For better generalization with reduced variance, the SMOTE-RUS-NC algorithm is incorporated into the framework of the RF classifier, producing the SRN-BRF classifier. This proposed novel ensemble classification algorithm outperformed the other popular ensemble techniques in almost all datasets. The performance measures obtained from this approach as well as other ensemble approaches are provided in Table 8 to 12.

**Table 8:** Performance comparison of the proposed ensemble algorithm with other ensemble techniques in terms of Accuracy (in percentage)

| Dataset | Imbalance Ratio | Over Bagging | SMOTE-Bagging | Over Boost | RUSBoost | Balanced Bagging | BRF | Easy Ensemble | SRN-BRF (Proposed) |
|---|---|---|---|---|---|---|---|---|---|
| wisconsin | 1.86 | 95.45 | 95.70 | 95.01 | 94.43 | 95.61 | 97.22 | 96.20 | 97.51 |
| yeast1 | 2.46 | 73.88 | 73.55 | 70.30 | 91.91 | 93.93 | 93.06 | 90.77 | 93.26 |
| vehicle1 | 2.90 | 78.44 | 78.38 | 76.82 | 72.68 | 76.24 | 77.30 | 77.30 | 76.95 |
| ecoli2 | 5.46 | 91.88 | 90.75 | 89.31 | 89.86 | 85.55 | 87.04 | 86.46 | 89.15 |
| yeast3 | 8.10 | 94.71 | 94.70 | 93.63 | 90.43 | 93.26 | 93.19 | 91.44 | 93.12 |
| ecoli3 | 8.60 | 90.45 | 90.39 | 89.34 | 88.62 | 87.45 | 85.91 | 85.02 | 87.12 |
| page-blocks0 | 8.79 | 96.01 | 95.50 | 92.91 | 92.09 | 95.47 | 93.51 | 93.00 | 93.62 |
| vowel0 | 9.98 | 96.39 | 96.13 | 95.90 | 95.65 | 96.06 | 95.76 | 96.46 | 97.17 |
| glass2 | 11.59 | 91.54 | 89.90 | 90.34 | 84.50 | 74.81 | 71.43 | 73.25 | 73.31 |
| glass4 | 15.47 | 87.00 | 75.88 | 89.37 | 92.42 | 85.26 | 83.35 | 87.66 | 94.78 |
| ecoli4 | 15.80 | 95.40 | 94.42 | 95.73 | 96.98 | 90.98 | 88.21 | 86.11 | 95.47 |
| abalone9-18 | 16.40 | 94.43 | 92.40 | 90.47 | 85.38 | 83.99 | 77.56 | 77.30 | 82.62 |
| flare-F | 23.79 | 80.24 | 81.11 | 80.30 | 85.83 | 76.80 | 79.17 | 69.41 | 84.23 |
| yeast4 | 28.10 | 96.49 | 96.12 | 91.78 | 87.52 | 87.06 | 82.20 | 81.53 | 86.05 |
| winequality-red-4 | 29.17 | 96.00 | 93.93 | 86.33 | 75.05 | 75.11 | 69.41 | 61.16 | 73.92 |
| yeast-1-2-8-9_vs_7 | 30.57 | 96.45 | 95.65 | 89.68 | 77.06 | 82.45 | 72.60 | 65.51 | 71.01 |
| yeast5 | 32.73 | 98.18 | 98.16 | 97.76 | 96.36 | 95.55 | 94.54 | 93.46 | 96.02 |
| ecoli-0-1-3-7_vs_2-6 | 39.14 | 96.18 | 0.00 | 95.00 | 79.08 | 79.08 | 69.90 | 64.80 | 84.69 |
| yeast6 | 41.40 | 98.15 | 98.03 | 95.95 | 92.25 | 91.51 | 90.42 | 86.51 | 89.55 |
| abalone-19_vs_10-11-12-13 | 49.69 | 97.92 | 96.88 | 90.69 | 83.35 | 80.88 | 66.90 | 65.92 | 77.36 |
| winequality-white-3-9_vs_5 | 58.28 | 98.12 | 97.95 | 94.82 | 83.63 | 83.51 | 74.09 | 67.49 | 87.98 |
| poker-8-9_vs_6 | 58.40 | 98.32 | 98.63 | 80.49 | 70.87 | 85.15 | 76.04 | 36.74 | 97.31 |
| winequality-red-3_vs_5 | 68.10 | 98.28 | 98.01 | 97.16 | 91.71 | 86.70 | 76.87 | 77.04 | 78.70 |
| abalone-20_vs_8-9-10 | 72.69 | 98.50 | 98.01 | 97.71 | 93.50 | 88.54 | 80.17 | 81.86 | 89.54 |
| poker-8_vs_6 | 85.88 | 98.85 | 99.04 | 87.22 | 71.67 | 78.19 | 55.82 | 33.71 | 94.58 |
| abalone19 | 129.44 | 99.20 | 98.65 | 91.45 | 86.21 | 80.17 | 68.62 | 68.96 | 76.97 |

**Table 9:** Performance comparison of the proposed ensemble algorithm with other ensemble techniques in terms of Sensitivity (in percentage)

| **Dataset** | **Imbalance Ratio** | **Over Bagging** | **SMOTE-Bagging** | **Over Boost** | **RUSBoost** | **Balanced Bagging** | **BRF** | **Easy Ensemble** | **SRN-BRF (Proposed)** |
|---|---|---|---|---|---|---|---|---|---|
| wisconsin | 1.86 | 91.94 | 92.74 | 92.57 | 90.80 | 92.90 | 98.32 | 97.48 | 99.57 |
| yeast1 | 2.46 | 39.30 | 49.20 | 69.96 | 81.40 | 90.15 | 92.57 | 91.91 | 94.41 |
| vehicle1 | 2.90 | 44.02 | 52.68 | 67.85 | 60.76 | 67.34 | 83.85 | 79.26 | 87.14 |
| ecoli2 | 5.46 | 76.24 | 78.67 | 86.47 | 81.00 | 86.67 | 88.67 | 90.33 | 88.67 |
| yeast3 | 8.10 | 71.60 | 76.52 | 87.19 | 69.67 | 90.15 | 93.20 | 91.32 | 94.41 |
| ecoli3 | 8.60 | 44.57 | 56.29 | 55.71 | 60.00 | 86.67 | 91.67 | 88.33 | 91.67 |
| page-blocks0 | 8.79 | 74.83 | 80.71 | 86.93 | 79.41 | 89.99 | 94.64 | 93.75 | 95.36 |
| vowel0 | 9.98 | 77.89 | 76.44 | 70.78 | 65.56 | 93.33 | 94.44 | 95.56 | 95.56 |
| glass2 | 11.59 | 11.67 | 15.24 | 27.67 | 55.00 | 40.00 | 90.00 | 90.00 | 85.00 |
| glass4 | 15.47 | 62.33 | 69.75 | 69.00 | 90.00 | 95.00 | 95.00 | 95.00 | 95.00 |
| ecoli4 | 15.80 | 67.50 | 72.50 | 71.00 | 90.00 | 85.00 | 95.00 | 95.00 | 95.00 |
| abalone9-18 | 16.40 | 15.92 | 31.72 | 42.19 | 45.50 | 55.00 | 72.50 | 75.50 | 75.00 |
| flare-F | 23.79 | 82.40 | 83.56 | 81.06 | 86.40 | 76.62 | 78.88 | 68.73 | 84.15 |
| yeast4 | 28.10 | 13.56 | 33.67 | 54.07 | 43.33 | 80.33 | 84.00 | 72.33 | 86.00 |
| winequality-red-4 | 29.17 | 99.22 | 96.73 | 88.15 | 76.27 | 75.55 | 69.66 | 61.00 | 74.44 |
| yeast-1-2-8-9_vs_7 | 30.57 | 7.00 | 10.00 | 33.00 | 36.67 | 46.67 | 73.33 | 66.67 | 73.33 |
| yeast5 | 32.73 | 60.89 | 72.36 | 77.08 | 84.50 | 91.00 | 98.00 | 98.00 | 98.00 |
| ecoli-0-1-3-7_vs_2-6 | 39.14 | 8.00 | 0.00 | 24.00 | 57.14 | 85.71 | 85.71 | 85.71 | 85.72 |
| yeast6 | 41.40 | 49.43 | 54.29 | 56.00 | 56.67 | 77.50 | 81.67 | 76.67 | 85.00 |
| abalone-19_vs_10-11-12-13 | 49.69 | 0.33 | 5.24 | 31.43 | 35.95 | 48.90 | 65.43 | 73.57 | 61.67 |
| winequality-white-3-9_vs_5 | 58.28 | 0.00 | 2.40 | 15.60 | 24.80 | 46.40 | 66.80 | 71.20 | 63.33 |
| poker-8-9_vs_6 | 58.40 | 0.00 | 18.40 | 24.00 | 23.60 | 45.60 | 82.00 | 51.20 | 93.33 |
| winequality-red-3_vs_5 | 68.10 | 0.00 | 0.00 | 2.00 | 12.00 | 42.00 | 52.00 | 47.00 | 70.00 |
| abalone-20_vs_8-9-10 | 72.69 | 8.47 | 28.60 | 48.27 | 61.60 | 73.87 | 88.33 | 82.13 | 85.00 |
| poker-8_vs_6 | 85.88 | 0.00 | 16.50 | 24.17 | 22.67 | 51.17 | 86.83 | 78.00 | 75.00 |
| abalone19 | 129.44 | 0.00 | 1.00 | 29.86 | 37.95 | 60.00 | 83.43 | 80.62 | 75.00 |

**Table 10:** Performance comparison of the proposed ensemble algorithm with other ensemble techniques in terms of Specificity (in percentage)

| Dataset | Imbalance Ratio | Over Bagging | SMOTE-Bagging | Over Boost | RUSBoost | Balanced Bagging | BRF | Easy Ensemble | SRN-BRF (Proposed) |
|---|---|---|---|---|---|---|---|---|---|
| wisconsin | 1.86 | 97.34 | 97.30 | 96.33 | 96.39 | 97.08 | 96.63 | 95.51 | 96.41 |
| yeast1 | 2.46 | 87.95 | 83.46 | 70.44 | 93.19 | 94.40 | 93.11 | 90.61 | 93.11 |
| vehicle1 | 2.90 | 90.32 | 87.25 | 79.92 | 76.78 | 79.33 | 75.04 | 76.62 | 73.44 |
| ecoli2 | 5.46 | 94.76 | 92.95 | 89.85 | 91.59 | 85.41 | 86.82 | 85.79 | 89.32 |
| yeast3 | 8.10 | 97.56 | 96.94 | 94.42 | 92.95 | 93.64 | 93.18 | 91.44 | 92.96 |
| ecoli3 | 8.60 | 95.80 | 94.37 | 93.27 | 92.00 | 87.67 | 85.33 | 84.67 | 86.67 |
| page-blocks0 | 8.79 | 98.41 | 97.18 | 93.58 | 93.53 | 96.09 | 93.39 | 92.92 | 93.43 |
| vowel0 | 9.98 | 98.24 | 98.10 | 98.42 | 98.67 | 96.33 | 95.89 | 96.56 | 97.33 |
| glass2 | 11.59 | 98.36 | 96.26 | 95.67 | 87.13 | 77.55 | 69.97 | 71.89 | 72.58 |
| glass4 | 15.47 | 88.65 | 76.39 | 90.80 | 93.00 | 85.00 | 83.00 | 87.50 | 95.00 |
| ecoli4 | 15.80 | 97.17 | 95.81 | 97.30 | 97.42 | 91.34 | 87.76 | 85.52 | 95.50 |
| abalone9-18 | 16.40 | 99.22 | 96.08 | 93.37 | 87.82 | 85.78 | 77.79 | 77.35 | 83.01 |
| flare-F | 23.79 | 27.69 | 22.81 | 62.03 | 72.00 | 81.00 | 86.00 | 86.00 | 86.00 |
| yeast4 | 28.10 | 99.45 | 98.34 | 93.13 | 89.10 | 87.29 | 82.13 | 81.85 | 86.04 |
| winequality-red-4 | 29.17 | 2.31 | 12.15 | 33.16 | 40.33 | 62.33 | 62.00 | 66.00 | 58.00 |
| yeast-1-2-8-9_vs_7 | 30.57 | 99.38 | 98.46 | 91.54 | 78.38 | 83.63 | 72.57 | 65.47 | 70.94 |
| yeast5 | 32.73 | 99.31 | 98.94 | 98.37 | 96.73 | 95.69 | 94.44 | 93.33 | 95.97 |
| ecoli-0-1-3-7_vs_2-6 | 39.14 | 98.33 | 0.00 | 96.52 | 79.89 | 78.84 | 69.31 | 64.02 | 84.66 |
| yeast6 | 41.40 | 99.33 | 99.09 | 96.92 | 93.09 | 91.85 | 90.60 | 86.74 | 89.64 |
| abalone-19_vs_10-11-12-13 | 49.69 | 99.82 | 98.67 | 91.83 | 84.26 | 81.50 | 66.93 | 65.77 | 77.67 |
| winequality-white-3-9_vs_5 | 58.28 | 99.80 | 99.59 | 96.18 | 84.64 | 84.15 | 74.21 | 67.43 | 88.47 |
| poker-8-9_vs_6 | 58.40 | 100.00 | 100.00 | 81.46 | 71.68 | 85.83 | 75.94 | 36.49 | 98.01 |
| winequality-red-3_vs_5 | 68.10 | 99.72 | 99.45 | 98.56 | 92.88 | 87.35 | 77.24 | 77.49 | 78.82 |
| abalone-20_vs_8-9-10 | 72.69 | 99.74 | 98.96 | 98.40 | 93.95 | 88.75 | 80.05 | 81.86 | 89.62 |
| poker-8_vs_6 | 85.88 | 100.00 | 100.00 | 87.96 | 72.23 | 78.51 | 55.47 | 33.22 | 94.79 |
| abalone19 | 129.44 | 99.97 | 99.41 | 91.93 | 86.59 | 80.33 | 68.51 | 68.88 | 76.99 |

**Table 11:** Performance comparison of the proposed ensemble algorithm with other ensemble techniques in terms of G-Mean (in percentage)

| **Dataset** | **Imbalance Ratio** | **Over Bagging** | **SMOTE-Bagging** | **Over Boost** | **RUSBoost** | **Balanced Bagging** | **BRF** | **Easy Ensemble** | **SRN-BRF (Proposed)** |
|---|---|---|---|---|---|---|---|---|---|
| wisconsin | 1.86 | 94.53 | 94.92 | 94.40 | 93.53 | 94.85 | 97.45 | 96.47 | **97.97** |
| yeast1 | 2.46 | 58.71 | 64.03 | 70.17 | 86.73 | 92.18 | 92.77 | 91.18 | **93.72** |
| vehicle1 | 2.90 | 62.86 | 67.69 | 73.50 | 67.68 | 72.82 | 79.24 | 77.77 | **79.86** |
| ecoli2 | 5.46 | 84.79 | 85.32 | **88.11** | 85.26 | 83.80 | 85.18 | 85.47 | 87.40 |
| yeast3 | 8.10 | 83.48 | 86.05 | 90.68 | 79.24 | 91.85 | 93.13 | 91.31 | **93.65** |
| ecoli3 | 8.60 | 64.42 | 72.37 | 71.33 | 68.43 | 86.52 | 88.12 | 86.11 | **88.85** |
| page-blocks0 | 8.79 | 85.63 | 88.42 | 90.00 | 85.79 | 92.73 | 93.78 | 93.08 | **94.22** |
| vowel0 | 9.98 | 86.84 | 86.10 | 83.05 | 77.26 | 94.45 | 94.93 | 95.88 | **96.31** |
| glass2 | 11.59 | 21.91 | 27.58 | 43.06 | 60.37 | 38.22 | 78.72 | 75.66 | **77.14** |
| glass4 | 15.47 | 72.69 | 70.14 | 78.35 | 90.51 | 87.14 | 86.77 | 90.49 | **94.50** |
| ecoli4 | 15.80 | 79.99 | 82.36 | 81.77 | 92.80 | 86.76 | 89.65 | 88.79 | **94.69** |
| abalone9-18 | 16.40 | 37.67 | 54.55 | 60.67 | 61.08 | 67.29 | 73.69 | 75.36 | **77.95** |
| flare-F | 23.79 | 31.31 | 36.74 | 58.61 | 77.99 | 78.12 | 82.03 | 75.31 | **84.77** |
| yeast4 | 28.10 | 31.22 | 56.19 | 70.72 | 61.24 | 83.08 | 82.08 | 75.93 | **85.20** |
| winequality-red-4 | 29.17 | 7.41 | 31.40 | 52.55 | 54.01 | **67.43** | 63.30 | 61.78 | 63.39 |
| yeast-1-2-8-9_vs_7 | 30.57 | 15.18 | 22.35 | 50.87 | 53.15 | 57.10 | 71.48 | 61.20 | **71.63** |
| yeast5 | 32.73 | 77.34 | 84.27 | 86.41 | 90.03 | 93.11 | 96.13 | 95.55 | **96.92** |
| ecoli-0-1-3-7_vs_2-6 | 39.14 | 10.99 | 0.00 | 29.30 | 47.12 | 73.51 | 67.75 | 63.38 | **77.43** |
| yeast6 | 41.40 | 68.43 | 71.91 | 71.81 | 67.73 | 83.15 | 85.12 | 80.44 | **86.87** |
| abalone-19_vs_10-11-12-13 | 49.69 | 0.82 | 12.35 | 46.31 | 49.81 | 59.69 | 65.09 | **67.33** | 67.04 |
| winequality-white-3-9_vs_5 | 58.28 | 0.00 | 5.37 | 30.13 | 37.73 | 59.69 | 68.84 | 67.45 | **72.97** |
| poker-8-9_vs_6 | 58.40 | 0.00 | 33.28 | 33.33 | 32.82 | 60.09 | 78.07 | 41.07 | **95.37** |
| winequality-red-3_vs_5 | 68.10 | 0.00 | 0.00 | 2.81 | 16.26 | 46.23 | 49.37 | 47.76 | **61.62** |
| abalone-20_vs_8-9-10 | 72.69 | 18.73 | 46.10 | 60.45 | 72.03 | 79.68 | 83.62 | 80.97 | **85.99** |
| poker-8_vs_6 | 85.88 | 0.00 | 25.73 | 35.68 | 31.07 | 57.87 | 68.98 | 50.20 | **79.03** |
| abalone19 | 129.44 | 0.00 | 2.43 | 50.79 | 53.60 | 68.16 | **75.09** | 73.67 | 74.85 |

**Table 12:** Performance comparison of the proposed ensemble algorithm with other ensemble techniques in terms of ROC-AUC (in percentage)

| **Dataset** | **Imbalance Ratio** | **Over Bagging** | **SMOTE-Bagging** | **Over Boost** | **RUSBoost** | **Balance Bagging** | **BRF** | **Easy Ensemble** | **SRN-BRF (Proposed)** |
|---|---|---|---|---|---|---|---|---|---|
| wisconsin | 1.86 | 94.64 | 95.02 | 94.45 | 93.60 | 94.99 | 97.47 | 96.49 | 97.99 |
| yeast1 | 2.46 | 63.63 | 66.33 | 70.20 | 87.29 | 92.27 | 92.84 | 91.26 | 93.76 |
| vehicle1 | 2.90 | 67.17 | 69.96 | 73.88 | 68.77 | 73.33 | 79.45 | 77.94 | 80.29 |
| ecoli2 | 5.46 | 85.50 | 85.81 | 88.16 | 86.29 | 86.04 | 87.74 | 88.06 | 89.00 |
| yeast3 | 8.10 | 84.58 | 86.73 | 90.80 | 81.31 | 91.89 | 93.19 | 91.38 | 93.68 |
| ecoli3 | 8.60 | 70.19 | 75.33 | 74.49 | 76.00 | 87.17 | 88.50 | 86.50 | 89.17 |
| page-blocks0 | 8.79 | 86.62 | 88.94 | 90.26 | 86.47 | 93.04 | 94.01 | 93.33 | 94.39 |
| vowel0 | 9.98 | 88.07 | 87.27 | 84.60 | 82.11 | 94.83 | 95.17 | 96.06 | 96.44 |
| glass2 | 11.59 | 55.01 | 55.75 | 61.67 | 71.07 | 58.78 | 79.99 | 80.95 | 78.79 |
| glass4 | 15.47 | 75.49 | 73.07 | 79.90 | 91.50 | 90.00 | 89.00 | 91.25 | 95.00 |
| ecoli4 | 15.80 | 82.34 | 84.15 | 84.15 | 93.71 | 88.17 | 91.38 | 90.26 | 95.25 |
| abalone9-18 | 16.40 | 57.57 | 63.90 | 67.78 | 66.66 | 70.39 | 75.14 | 76.42 | 79.01 |
| flare-F | 23.79 | 55.05 | 53.18 | 71.54 | 79.20 | 78.81 | 82.44 | 77.36 | 85.08 |
| yeast4 | 28.10 | 56.51 | 66.01 | 73.60 | 66.22 | 83.81 | 83.06 | 77.09 | 86.02 |
| winequality-red-4 | 29.17 | 50.76 | 54.44 | 60.66 | 58.30 | 68.94 | 65.83 | 63.50 | 66.22 |
| yeast-1-2-8-9_vs_7 | 30.57 | 53.19 | 54.23 | 62.27 | 57.53 | 65.15 | 72.95 | 66.07 | 72.13 |
| yeast5 | 32.73 | 80.10 | 85.65 | 87.73 | 90.62 | 93.35 | 96.22 | 95.66 | 96.99 |
| ecoli-0-1-3-7_vs_2-6 | 39.14 | 53.17 | 0.00 | 60.26 | 68.52 | 82.28 | 77.51 | 74.87 | 85.19 |
| yeast6 | 41.40 | 74.38 | 76.69 | 76.46 | 74.88 | 84.67 | 86.13 | 81.70 | 87.31 |
| abalone-19_vs_10-11-12-13 | 49.69 | 50.08 | 51.95 | 61.63 | 60.11 | 65.20 | 66.18 | 69.67 | 69.67 |
| winequality-white-3-9_vs_5 | 58.28 | 49.90 | 50.99 | 55.89 | 54.72 | 65.27 | 70.51 | 69.31 | 75.90 |
| poker-8-9_vs_6 | 58.40 | 50.00 | 59.20 | 52.73 | 47.64 | 65.72 | 78.97 | 43.85 | 95.67 |
| winequality-red-3_vs_5 | 68.10 | 49.86 | 49.72 | 50.28 | 52.44 | 64.68 | 64.62 | 62.24 | 74.41 |
| abalone-20_vs_8-9-10 | 72.69 | 54.10 | 63.78 | 73.33 | 77.77 | 81.31 | 84.19 | 81.99 | 87.31 |
| poker-8_vs_6 | 85.88 | 50.00 | 58.25 | 56.06 | 47.45 | 64.84 | 71.15 | 55.61 | 84.90 |
| abalone19 | 129.44 | 49.99 | 50.20 | 60.90 | 62.27 | 70.17 | 75.97 | 74.75 | 76.00 |

SRN-BRF is a modification of the original BRF classifier, and as can be observed from the results, the proposed algorithm outperformed the BRF approach in almost all datasets (25 out of 26). Although the improvement is small in some cases, in several other cases, the improvement is quite significant compared to BRF. As for some of the other ensemble approaches like Over-Bagging, SMOTE-Bagging, Over-Boost, or RUSBoost, the performance of these ensemble algorithms is significantly lower. BB, BRF, and Easy Ensemble provided comparatively better results. Nonetheless, our proposed approach is able to outperform all of them (including all the other sampling techniques), in several cases by a large margin. For instance, in the 'poker-8_vs_6' dataset, the g-mean score obtained from the proposed SRN-BRF classifier is 79.03%, which is considerably higher than other approaches like Easy Ensemble (50.20%), BRF (68.98%), or BB (57.87%). Algorithms like Over-Bagging, SMOTE-Bagging, or RUSBoost provided relatively poor performance – 0%, 25.73%, and 31.07%, respectively. The base algorithm SMOTE-RUS-NC achieved a 75.32% g-mean score, which is higher than other ensemble techniques but lower than the SRN-BRF classifier. The ensemble approach provided better results on average, as compared to the base sampling technique SMOTE-RUS-NC. The main drawback of the ensemble techniques is that they are more time-consuming. Compared to that, the sampling techniques work as a data preprocessing approach before training a classification algorithm, which makes them relatively fast.

The difference in performance (based on G-Mean score) among these ensemble approaches on some highly imbalanced datasets is illustrated in Figure 4.

On the basis of the acquired findings, it can be inferred that both of the proposed algorithms are quite effective in handling imbalanced data. They were able to surpass other cutting-edge sampling methods and can therefore be extremely useful in imbalanced domains. Instead of relying on a single sampling technique to balance the class distribution, the study demonstrates that a proper hybridization between different sampling approaches can be proven more effective in addressing the imbalanced learning problem. To further improve the performance and reduce the variance, the hybrid sampling strategy can be integrated with the ensemble framework, producing a more powerful ensemble algorithm for imbalanced learning.

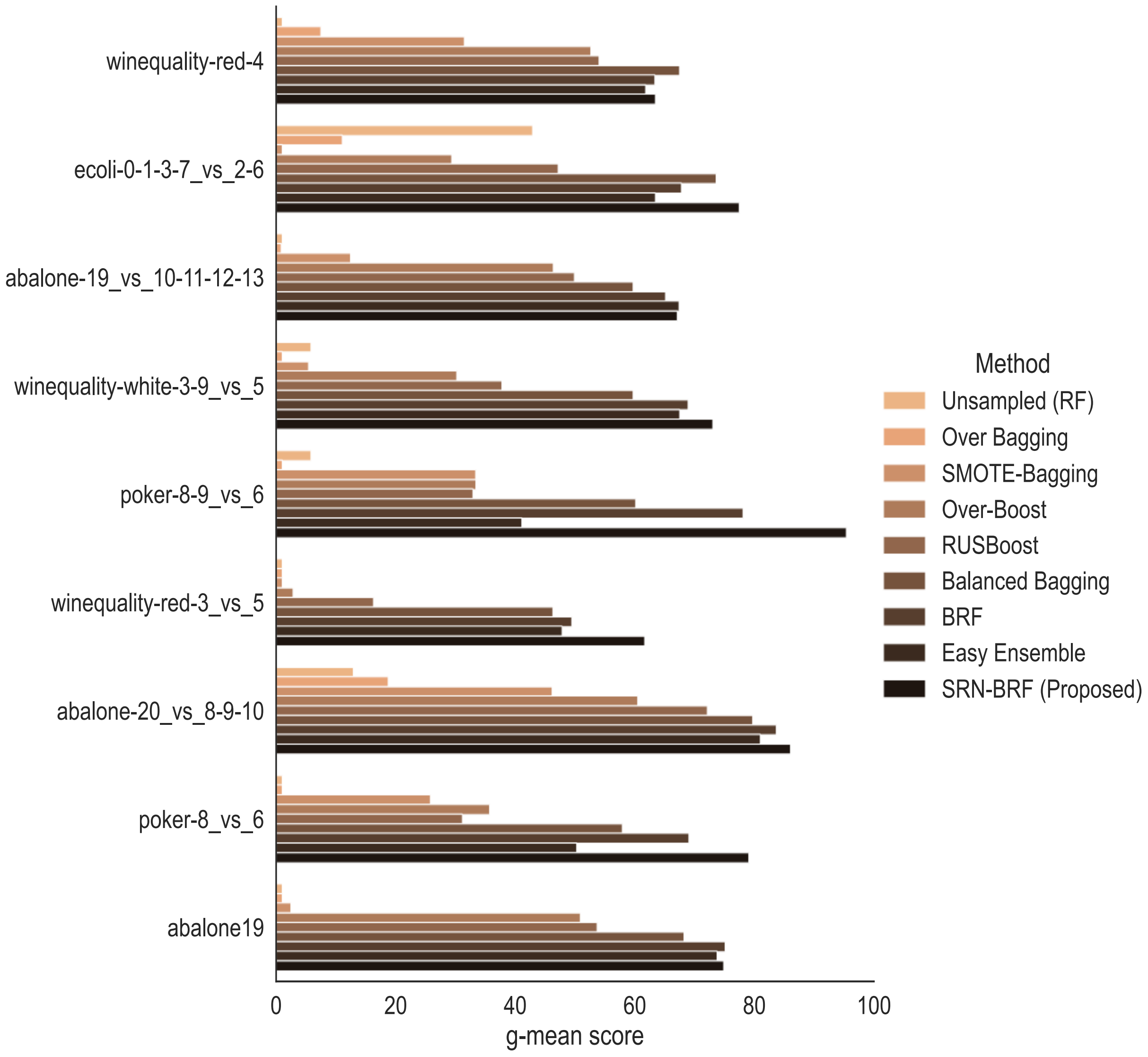

**Figure 4.** Performance comparison (in terms of G-Mean score) of the proposed SRN-BRF approach with other ensemble algorithms

## 5. Conclusion

Data sampling is an indispensable step when dealing with imbalanced data. Standard classification algorithms cannot perform well if one class is underrepresented. Oversampling, undersampling, or hybridization between the two approaches can be utilized to minimize the disparity between the classes and train the classifier on that balanced data. SMOTE, Safe-level-SMOTE, ADASYN, Borderline-SMOTE, CNN, RUS, IHT, SMOTE-ENN, RUSBoost, BRF, Balanced Bagging, etc. are some of the state-of-the-art sampling techniques. They are usually capable of improving the overall performance to some extent. However, these techniques have certain limitations and do not work well in all cases. For instance, in high imbalance scenarios, performance improvement using these algorithms is quite limited. In some cases, the algorithms failed utterly to counter bias. The CNN method is quite time-consuming and becomes infeasible on large datasets. The RUS method, on the other hand, suffers from information loss and high variance. SMOTE and its variations are impractical when the number of minority class samples is limited.

To overcome the limitations of these techniques and to develop a sampling methodology capable of providing consistent performance on a wide variety of imbalanced datasets, a novel hybrid sampling algorithm has been proposed in this manuscript. A sophisticated framework to combine three sampling algorithms (SMOTE, NC, and RUS) has been developed. The proposed SMOTE-RUS-NC algorithm is further incorporated into the ensemble learning framework. This ensemble approach, SRN-BRF, achieves better performance and generalization with reduced variance. The performance of the proposed techniques has been validated on 26 publicly available imbalanced datasets. The results have been compared with other popular sampling techniques in the imbalanced domain. The proposed algorithms outperformed other state-of-the-art methods in almost all datasets, in some cases by a large margin. Especially in highly imbalanced datasets, the proposed algorithms achieved unparalleled performance. This signifies the superiority of the proposed techniques and their potential to be an effective sampling approach in imbalanced classification tasks.

There are several future research scopes. Different variations of the SMOTE algorithm have been introduced by researchers over the years. They might be proven more effective compared to the original SMOTE algorithm utilized in this study. The NC algorithm has been used initially to reduce the imbalance ratio. Other heuristic undersampling approaches that can provide a greater reduction in class imbalance can also be used in its place. Moreover, this study focuses on the binary classification task. The same idea can be extended to multi-class classification tasks as well.

In conclusion, we believe that our proposed methodologies can handle imbalanced data more effectively than current state-of-the-art approaches in the imbalanced domain. The hybridization framework suggested in this study can be explored further to obtain an even more robust algorithm. These can be particularly valuable in the imbalanced learning domain.

# 6. Conflicts of Interest Disclosure

The authors declare that there is no conflict of interest.

# 7. Sources of Funding

This research did not receive any specific grant from funding agencies in the public, commercial, or not-for-profit sectors.

## References

[1] Chawla NV, Bowyer KW, Hall LO, Kegelmeyer WP. SMOTE: Synthetic minority over-sampling technique. J Artif Intell Res 2002;16:321–57. https://doi.org/10.1613/jair.953

[2] Fernandez A, Garcia S, Herrera F, Chawla NV. SMOTE for learning from imbalanced data: Progress and challenges, marking the 15-year anniversary. J Artif Intell Res 2018;61:863–905. https://doi.org/10.1613/jair.1.11192

[3] He H, Bai Y, Garcia EA, Li S. ADASYN: Adaptive synthetic sampling approach for imbalanced learning. 2008 IEEE International Joint Conference on Neural Networks (IEEE World Congress on Computational Intelligence), IEEE; 2008. https://doi.org/10.1109/IJCNN.2008.4633969

[4] Han H, Wang W-Y, Mao B-H. Borderline-SMOTE: A new over-sampling method in imbalanced data sets learning. Lecture Notes in Computer Science, Berlin, Heidelberg: Springer Berlin Heidelberg; 2005, p. 878–87. https://doi.org/10.1007/11538059_91

[5] Bunkhumpornpat C, Sinapiromsaran K, Lursinsap C. DBSMOTE: Density-based synthetic minority over-sampling TEchnique. Appl Intell 2012;36:664–84. https://doi.org/10.1007/s10489-011-0287-y

[6] Bunkhumpornpat C, Sinapiromsaran K, Lursinsap C. Safe-level-SMOTE: Safe-level-synthetic minority over-sampling TEchnique for handling the class imbalanced problem. Advances in Knowledge Discovery and Data Mining, Berlin, Heidelberg: Springer Berlin Heidelberg; 2009, p. 475–82. https://doi.org/10.1007/978-3-642-01307-2_43

[7] Two modifications of CNN. IEEE Trans Syst Man Cybern 1976;SMC-6:769–72. https://doi.org/10.1109/TSMC.1976.4309452

[8] Hart P. The condensed nearest neighbor rule (Corresp.). IEEE Trans Inf Theory 1968;14:515–6. https://doi.org/10.1109/TIT.1968.1054155

[9] Kubat M, Holte R, Matwin S. Learning when negative examples abound. Machine Learning: ECML-97, Berlin, Heidelberg: Springer Berlin Heidelberg; 1997, p. 146–53. https://doi.org/10.1007/3-540-62858-4_79

[10] Zhang J, Mani I. KNN Approach to Unbalanced Data Distributions: A Case Study Involving Information Extraction. Proceedings of the ICML'2003 Workshop on Learning from Imbalanced Datasets, 2003

[11] Newaz A, Muhtadi S, Haq FS. An intelligent decision support system for the accurate diagnosis of cervical cancer. Knowl Based Syst 2022;245:108634. https://doi.org/10.1016/j.knosys.2022.108634

[12] Batista GEAPA, Bazzan ALC, Monard MC. Balancing training data for automated annotation of keywords: A case study. UfrgsBr n.d. https://www.inf.ufrgs.br/maslab/pergamus/pubs/balancing-training-data-for.pdf (accessed August 6, 2022).

[13] Batista GEAPA, Prati RC, Monard MC. A study of the behavior of several methods for balancing machine learning training data. SIGKDD Explor 2004;6:20–9. https://doi.org/10.1145/1007730.1007735

[14] Blagus R, Lusa L. SMOTE for high-dimensional class-imbalanced data. BMC Bioinformatics 2013;14:106. https://doi.org/10.1186/1471-2105-14-106

[15] Wilson DL. Asymptotic properties of nearest neighbor rules using edited data. IEEE Trans Syst Man Cybern 1972;SMC-2:408–21. https://doi.org/10.1109/TSMC.1972.4309137

[16] Ma L, Fan S. CURE-SMOTE algorithm and hybrid algorithm for feature selection and parameter optimization based on random forests. BMC Bioinformatics 2017;18:169. https://doi.org/10.1186/s12859-017-1578-z

[17] Barua S, Islam MM, Yao X, Murase K. MWMOTE--majority weighted minority oversampling technique for imbalanced data set learning. IEEE Trans Knowl Data Eng 2014;26:405–25. https://doi.org/10.1109/TKDE.2012.232

[18] Tahir MA, Kittler J, Yan F. Inverse random under sampling for class imbalance problem and its application to multi-label classification. Pattern Recognit 2012;45:3738–50. https://doi.org/10.1016/j.patcog.2012.03.014

[19] Kim H-J, Jo N-O, Shin K-S. Optimization of cluster-based evolutionary undersampling for the artificial neural networks in corporate bankruptcy prediction. Expert Syst Appl 2016;59:226–34. https://doi.org/10.1016/j.eswa.2016.04.027

[20] Yu H, Ni J, Zhao J. ACOSampling: An ant colony optimization-based undersampling method for classifying imbalanced DNA microarray data. Neurocomputing 2013;101:309–18. https://doi.org/10.1016/j.neucom.2012.08.018

[21] Díez-Pastor JF, Rodríguez JJ, García-Osorio C, Kuncheva LI. Random Balance: Ensembles of variable priors classifiers for imbalanced data. Knowl Based Syst 2015;85:96–111. https://doi.org/10.1016/j.knosys.2015.04.022

[22] Seiffert C, Khoshgoftaar TM, Van Hulse J, Napolitano A. RUSBoost: A hybrid approach to alleviating class imbalance. IEEE Trans Syst Man Cybern A Syst Hum 2010;40:185–97. https://doi.org/10.1109/TSMCA.2009.2029559

[23] Liu X-Y, Wu J, Zhou Z-H. Exploratory undersampling for class-imbalance learning. IEEE Trans Syst Man Cybern B Cybern 2009;39:539–50. https://doi.org/10.1109/TSMCB.2008.2007853

[24] Chen C, Liaw A. Using random forest to learn imbalanced data. Berkeley.edu n.d. https://statistics.berkeley.edu/sites/default/files/tech-reports/666.pdf (accessed August 6, 2022)

[25] Wang S, Yao X. Diversity analysis on imbalanced data sets by using ensemble models. 2009 IEEE Symposium on Computational Intelligence and Data Mining, IEEE; 2009, p. 324–31. https://doi.org/10.1109/CIDM.2009.4938667

[26] Xu-Ying Liu, Jianxin Wu, Zhi-Hua Zhou. Exploratory Undersampling for Class-Imbalance Learning. IEEE Transactions on Systems, Man, and Cybernetics, Part B (Cybernetics) 2009;39:539–50. https://doi.org/10.1109/TSMCB.2008.2007853

[27] Nanni L, Fantozzi C, Lazzarini N. Coupling different methods for overcoming the class imbalance problem. Neurocomputing 2015;158:48–61. https://doi.org/10.1016/j.neucom.2015.01.068

[28] 8. Common pitfalls and recommended practices — Version 0.9.1 n.d. https://imbalanced-learn.org/stable/common_pitfalls.html (accessed August 6, 2022)

[29] KEEL: A software tool to assess evolutionary algorithms for Data Mining problems (regression, classification, clustering, pattern mining and so on). Ugr.es n.d. https://sci2s.ugr.es/keel/imbalanced.php (accessed August 6, 2022)

[30] UCI Machine Learning Repository. Uci.edu n.d. https://archive.ics.uci.edu/ml/index.php (accessed August 6, 2022)